\documentclass[journal]{IEEEtran}
\usepackage{amsmath,amssymb,amsfonts}
\usepackage{cases}
\usepackage{color,xcolor}
\usepackage{graphicx}
\usepackage{subfigure}
\usepackage{algorithm}
\usepackage{algorithmic}
\usepackage{epstopdf}
\usepackage{multirow}
\usepackage{rotating}
\usepackage{booktabs}
\usepackage{tikz}
\usepackage{newtxmath}
\usepackage{acronym}
\usepackage{epstopdf}
\usepackage{xurl}
\usepackage{pifont}
\usepackage[demo,
            export]{adjustbox}  
\usepackage{tabularray}
\usepackage[table]{xcolor}
\usepackage{colortbl}
\usepackage[colorlinks,
linkcolor=blue,
anchorcolor=blue,
citecolor=blue
]{hyperref}
\usepackage{amssymb}
\usepackage[colorlinks,linkcolor=blue]{hyperref}
\usepackage{array} 
\usepackage{catchfilebetweentags}
\definecolor{tableTitleColor}{rgb}{0.8,0.8,0.8}
\begin{document} 

\title{PICANet: Physics-Informed Cascaded Asymmetric Network for Infrared Small Target Detection}
	
\author{Jingjing Liu, \IEEEmembership{Member,~IEEE}, Yinchao Han, Xianchao Xiu, \IEEEmembership{Member,~IEEE}, Jianhua Zhang,\\ \IEEEmembership{Senior Member,~IEEE}, and Wanquan Liu, \IEEEmembership{Senior Member,~IEEE}

\thanks{This work was supported in part by the National Natural Science Foundation of China under Grant 62204044 and Grant 12371306,  and in part by the State Key Laboratory of Integrated Chips and Systems under Grant SKLICS-K202302. (\textit{Corresponding author: Xianchao Xiu}.)}
\thanks{J. Liu, Y. Han, and J. Zhang are with the Shanghai Key Laboratory of Automobile Intelligent Network Interaction Chip and System,  School of Microelectronics,  Shanghai University, Shanghai 200444,  China (e-mail: \{jjliu, xiaobing252924, jhzhang\}@shu.edu.cn).}
\thanks{X. Xiu is with the School of Mechatronic Engineering and Automation,  Shanghai University,  Shanghai 200444, China (e-mail: xcxiu@shu.edu.cn).}
\thanks{W. Liu is with the School of Intelligent Systems Engineering, Sun Yat-sen University, Guangzhou 510275, China (e-mail: liuwq63@mail.sysu.edu.cn).}
}

\maketitle	
\begin{abstract}
Infrared small target detection (ISTD) is an important research direction in image  processing. However, existing  methods are limited by severe background noise propagation and target degradation in high-level semantic features. To address these limitations, this paper proposes a \textit{plug-and-play} physics-informed cascaded asymmetric network, named PICANet. Specifically, we construct a hierarchical prior decoupling module to explicitly extract low-level and high-level physical information, thereby characterizing target features at different levels rather than relying solely on convolutional extraction. Furthermore, a dual-prior interactive fusion module is developed to  dynamically  refine target representations while suppressing complex background clutter. Unlike previous work, a multi-level cross-feature attention module with the cascaded asymmetric mechanism is introduced to achieve precise alignment between high-level semantics and low-level spatial details. Extensive experiments demonstrate that the proposed PICANet outperforms state-of-the-art ISTD methods, showing satisfactory detection accuracy even against complex backgrounds. Our code is available at \url{https://github.com/xianchaoxiu/PICANet}.
\end{abstract}


\IEEEpeerreviewmaketitle

\section{Introduction}
\IEEEPARstart{I}{nfrared} small target detection (ISTD) has gained critical prominence in many fields such as remote sensing \cite{lu2026hyperistd}, intelligent transportation \cite{liu2025enhancing}, aerospace systems \cite{li2025ilnet}, and medical imaging \cite{liu2025graph}. Nevertheless, ISTD remains a highly challenging task, owing to several formidable obstacles inherent to the infrared imaging process \cite{kumar2025small}. For example, the extremely low signal-to-noise ratio (SNR) significantly limits the ability to obtain texture information \cite{sui2026samamba}, and environmental variations frequently cause random shape fluctuations, hindering accurate capture of target contours \cite{deng2025cspenet}. Consequently, ISTD has attracted widespread attention from both academia and industry. Interested readers can refer to \cite{yang2025deep, hua2025survey}.

Over the past few decades, existing  ISTD methods can be broadly categorized into three paradigms: \textit{model-driven} \cite{zhou2022infrared, liu2024infraredb, yin2026dnn}, \textit{data-driven} \cite{wu2023uiu, liu2024infrared, li2026dynamic}, and \textit{model-data-driven} \cite{wu2024rpcanet, liu2025ctvnet, liu2025lightweight}. In particular, \textit{model-driven} methods exhibit favorable physical interpretability, as they are established on rigorous mathematical formulations that explicitly characterize target and background priors. In contrast, \textit{data-driven} methods leverage convolutional operations and parallel computing capabilities to  achieve pixel-level segmentation accuracy. By inheriting the merits of the above two paradigms, \textit{model-data-driven} methods retain superior interpretability and powerful feature representation capability. We will discuss these three categories below.

\textit{Model-driven} ISTD methods exploit the distinct statistical characteristics of targets and backgrounds, and can be divided into three categories. The first category refers to background suppression-based methods, which isolate targets by estimating and filtering out background components \cite{bai2010infrared, zhao2020infrared}. The second category originates from the human visual system (HVS), which enhances salient target regions by calculating local contrast metrics \cite{wei2016multiscale, bai2018derivative}. The third category relies on low-rank and sparse decomposition, which leverages the low-rank property of infrared backgrounds and the intrinsic sparsity of small targets for target-background decoupling \cite{zhang2019infrared, kong2021infrared}. 
Notably, these methods possess clear physical interpretability and do not require network training. However, when dealing with multi-scale targets and complex backgrounds, their inherent mathematical models suffer from high parameter sensitivity and limited generalization capability \cite{luo2024revisiting,liu2025star}.

\begin{figure*}[t]
    \centering
    \includegraphics[width = 0.99\textwidth]{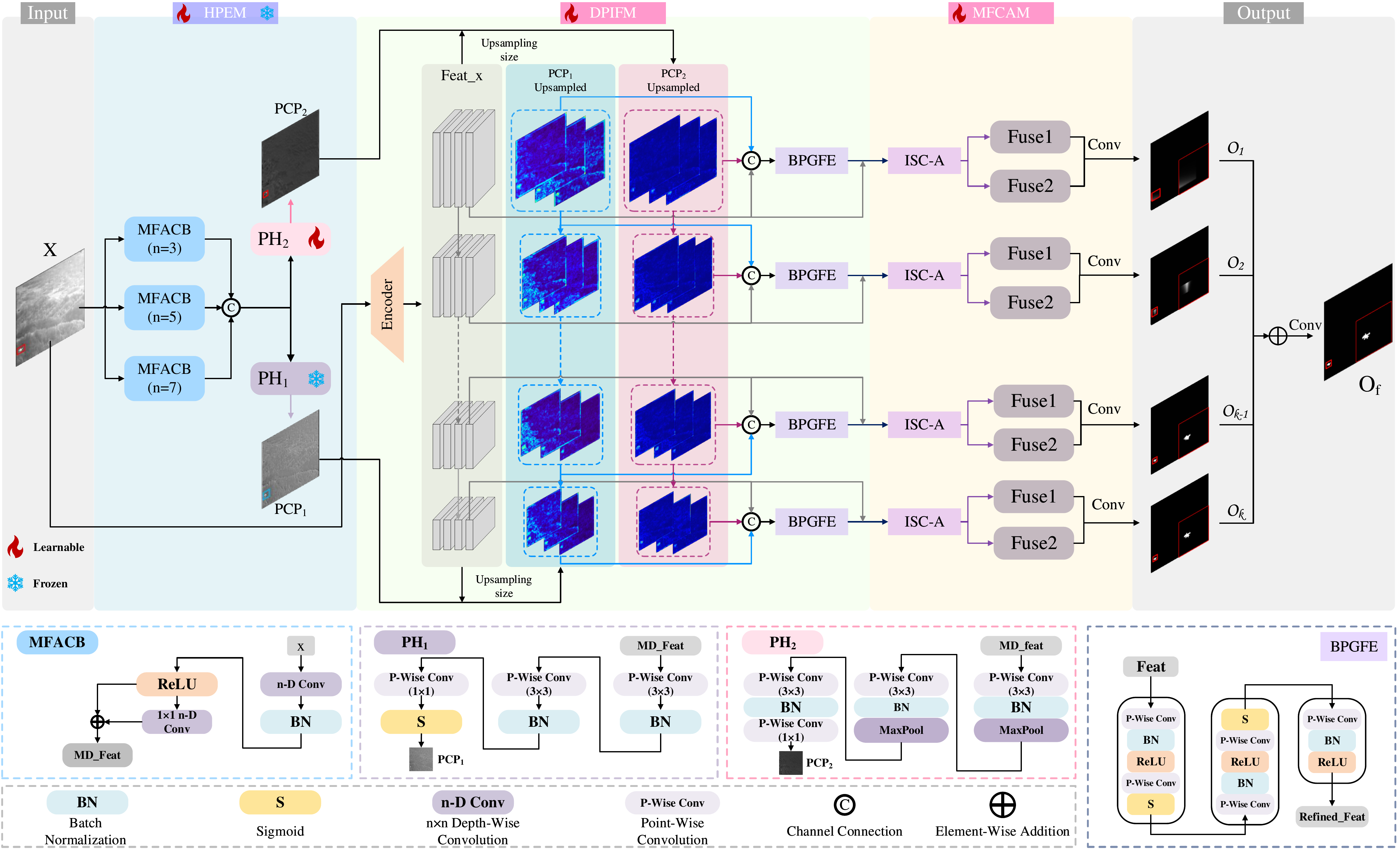} 
    \caption{Overall architecture of our proposed PICANet. It mainly consists of three core components: (1) HPDM extracts geometric and semantic physical priors from raw input images; (2) DPIFM injects dual prior information into multi-scale backbone features to strengthen target feature representations; (3) MCFAM aligns and fuses multi-level feature hierarchies via an interactive spatial-channel attention mechanism to generate high-fidelity target prediction maps.} 
    \label{prior_extract_fuse_flow} 
\end{figure*}

\textit{Data-driven} ISTD methods have achieved remarkable pixel-level accuracy thanks to the emergence of high-quality datasets and advanced computing resources. Their detection performance primarily stems from end-to-end semantic backbone networks that extract multi-level target features through multi-level pooling. One approach is the U-Net-based semantic segmentation framework, which uses encoder-decoder structures and skip connections to fuse spatial details with semantic contexts \cite{chen2022local, zhang2024irprunedet}. Another approach relies on the ResNet-style feature extractor, where stacked residual blocks capture highly abstract semantics for target localization \cite{sun2023receptive, wu2024saliency}. However, these backbones have some limitations for the ISTD task. For example, U-Net unconditionally concatenates low-level spatial features with high-level semantic features, lacking deterministic constraints or prior guidance \cite{liu2025lightweight}, while ResNet depends on excessive oversampling, inevitably suppressing fine contour features of small targets  \cite{yuan2026seeing}. Recently, due to the powerful long-range dependency modeling and global feature interaction capabilities,
Transformer-based ISTD methods have become a compelling alternative and achieve superior target extraction and robust background suppression; see, e.g., \cite{liu2023infrared, qin2025osformer, jiang2026bidirectional}.

\textit{Model-data-driven} ISTD methods have been developed to circumvent the inherent black-box drawback of the above \textit{data-driven} methods. Some studies integrate handcrafted priors derived from HVS into deep learning pipelines by calculating local contrast features to generate saliency maps \cite{dai2021attentional, yu2022infrared}. However, such manually designed priors embedded in the high-level space frequently cause mismatch between spatial and semantic features \cite{wang2025paying}. 
Another way is to unroll mathematical models into deep networks \cite{xiong2025drpca,wu2026rpcanet}, which is called deep unfolding or algorithm unfolding. This integrated scheme simultaneously inherits the  explicit priors of \textit{model-driven} methods and exploits the adaptive optimization superiority of \textit{data-driven} methods via end-to-end learning \cite{wu2024rpcanet, liu2025ctvnet}. It should be noted that these models may limit the detection performance because a single mathematical paradigm struggles to address the complex relationships between target contours, spatial information, and semantic information \cite{liu2025lightweight}.

Motivated by these observations, we propose a plug-and-play physics-informed cascaded asymmetric network (PICANet) for ISTD, as shown in Fig. \ref{prior_extract_fuse_flow}. In particular, the prior decoupling and bidirectional prior-guided fusion mechanisms embedded in PICANet can be flexibly integrated with arbitrary network architectures. This design resolves the misalignment problem of different-level target information in different backbones. To fully validate its effectiveness and generalization, we instantiate the proposed PICANet with U-Net and ResNet to construct two variants, i.e.,  PICANet (U) and PICANet (F). Extensive experimental results illustrate their prominent detection performance against existing ISTD competitors. In summary, the main contributions of this paper are as follows:

\begin{itemize}
 \item We construct a novel hierarchical prior decoupling module (HPDM) built upon frozen multi-directional feature aggregation calculation blocks. Unlike saliency-based priors, this module decouples the raw input into two physical constraint priors (PCPs), i.e., low-level geometric prior ($\mathrm{PCP}_1$) and high-level semantic prior ($\mathrm{PCP}_2$).

\item We introduce an effective dual prior interactive fusion module (DPIFM) that utilizes extracted physical priors as local spatial and semantic gates to actively inject deterministic constraints into the deep learning hierarchy, thus optimizing the backbone representations. This mechanism dynamically sharpens the features of small spatial targets while suppressing background clutter.

\item We develop an interactive  multi-level cross-feature attention module (MCFAM) through the cascaded asymmetric mechanism to achieve cross-layer fusion. This design resolves the misalignment between high-level semantic content and low-level spatial structure in  networks.
\end{itemize}

The rest of this paper is organized as follows. Section \ref{Related} gives some related work. Section \ref{Methodology} details the proposed PICANet. Section \ref{Experiment} provides the detection results and analysis. Section \ref{Conclusion} concludes this paper with future directions.

\begin{figure*}[t]
    \centering
    \includegraphics[width = 0.95\textwidth]{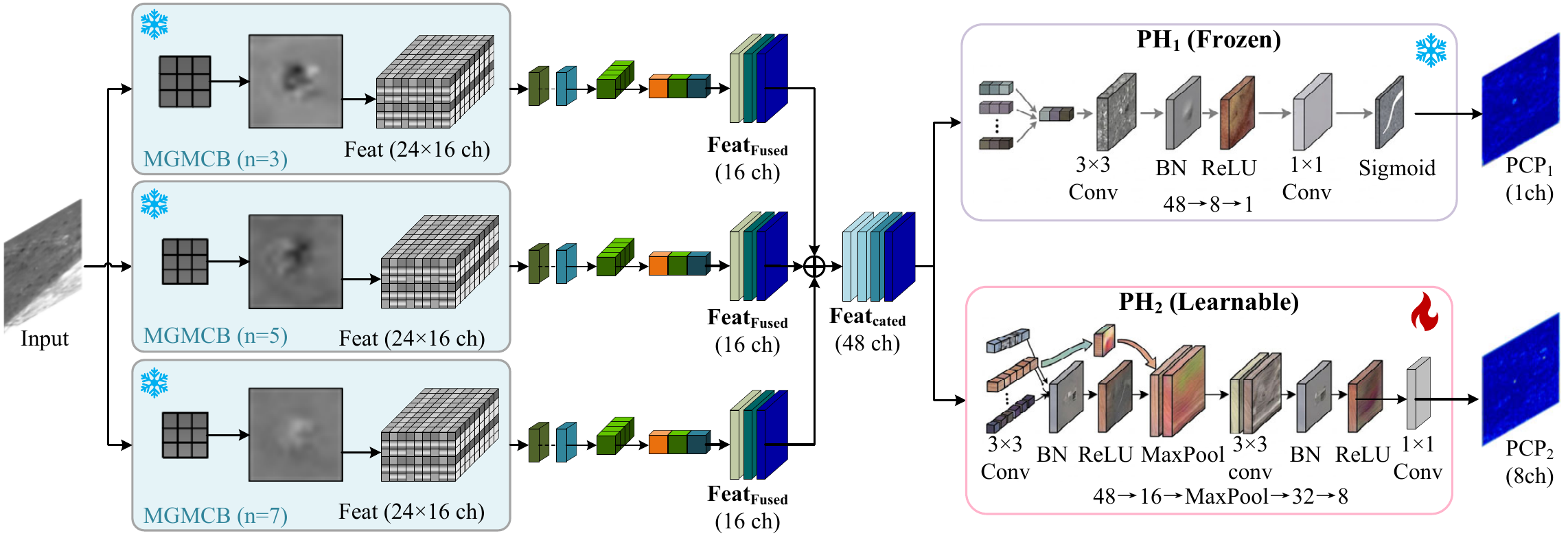} 
    \caption{Detailed architecture of HPDM. It consists of multi-directional feature aggregation calculation blocks (MFACBs) and dual-stream prior heads ($\mathrm{PH}_1$ and $\mathrm{PH}_2$) to decouple spatial and semantic priors.} 
    \label{fig:HPDM} 
\end{figure*}

\section{Related Work}\label{Related}

\subsection{Physics-Informed Networks}

Physics-informed networks can obtain reliable representations of small infrared targets by considering  local contrast, target sparsity, contour structure, and background smoothness. As an early representative effort, Dai et al. \cite{dai2021attentional} proposed an attention-based method called ALCNet by combining explicit target-background contrast modeling with end-to-end representation learning. Subsequently, Zhang et al. \cite{zhang2022isnet} developed ISNet by introducing a Taylor finite-difference-inspired edge block and a bidirectional attention aggregation module. Recently, Li et al. \cite{li2024usesnet} proposed USES-Net through embedding patch-level contrast and spatial-distribution knowledge into a Transformer-based encoder-decoder. Deng et al. \cite{deng2025cspenet} developed CSPENet by exploiting the convergent gradient patterns around small targets to construct complementary priors. Hao et al. \cite{hao2026deep} developed DURD as a data-driven residual decomposition network.

Deep unfolding is another commonly used technique for integrating physical priors. Wu et al. \cite{wu2024rpcanet} proposed RPCANet by unfolding robust principal component analysis (RPCA) to estimate the low-rank background and sparse target. Building on this, Xiong et al. \cite{xiong2025drpca} introduced DRPCA-Net, which employs an input-conditioned hypernetwork to dynamically generate iteration-wise parameters. Liu et al. \cite{liu2025ctvnet} proposed CTVNet by incorporating correlated total variation and local gradient smoothness in an unfolding network. 
Although these methods generally offer better interpretability and more constrained feature learning than black-box architectures, their effectiveness still depends on the validity of the physical priors. Therefore, if actual conditions deviate from the preset assumptions, missed detections or residual clutter may occur.

\subsection{Attention Mechanism}

Because small infrared targets typically occupy only a few pixels, they are easily overwhelmed by high-intensity clutter. Attention mechanisms can adaptively reweight spatial locations, feature channels, or contextual dependencies, enabling networks to allocate greater representational capacity to target-related cues. For example, Ren et al. \cite{ren2021dnanet} developed DNANet, where densely connected nested modules promote repeated interaction between shallow spatial details and deep semantic features. Tong et al. \cite{tong2021eaau} introduced an enhanced asymmetric attention U-Net to model directional contextual dependencies, while Zhang et al. \cite{zhang2021agpcnet} proposed AGPCNet to aggregate multi-scale contextual information under attention-based guidance.

Despite these advances, most attention-based architectures still heavily rely on repeated down-sampling and up-sampling to aggregate multi-level features. For extremely small targets, down-sampling may irreversibly weaken target responses, whereas subsequent up-sampling cannot fully recover the lost spatial details. Furthermore, features from different resolutions often exhibit inconsistent spatial locations, receptive fields, and semantic distributions. These limitations motivate us to develop attention mechanisms that can maintain spatial correspondences during resolution variations and explicitly align multi-level features before adaptive enhancement and fusion.

\begin{figure*}[t]
    \centering
    \includegraphics[width = 0.90\textwidth]{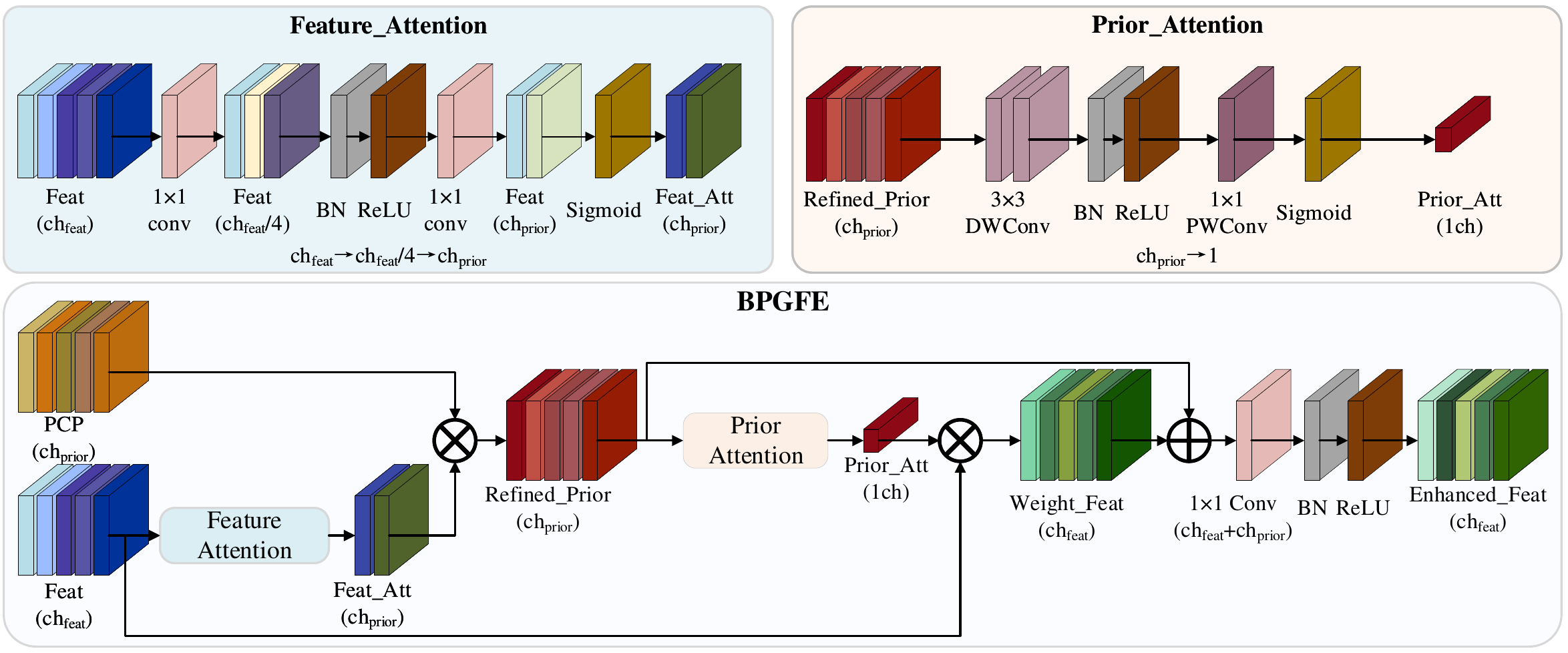} 
    \caption{Detailed structure of BPGFE. It adaptively modulates deep semantic features using the decoupled PCPs through a dual-attention mechanism within each level of the backbones.} 
    \label{fig:BPGFE} 
\end{figure*}

\section{Methodology}\label{Methodology}

This section first describes the detailed structures of HPDM, DPIFM, and MCFAM in Fig. \ref{prior_extract_fuse_flow}, then gives the loss function, and finally explains the plug-and-play property.

\subsection{Hierarchical Prior Decoupling Module (HPDM)}\label{HPDM}

The proposed HPDM shifts from simple local differences to physical modeling, thus addressing the bottleneck problem of high false alarm rates for background elements with similar local contrast \cite{yu2020searching, wu2023repisd}. As illustrated in Fig. \ref{fig:HPDM}, the module utilizes frozen multi-directional feature aggregation calculation blocks to explicitly quantify the centrally symmetric Gaussian morphology of valid targets. The freezing of mathematical operators preserves a consistent physical representation, thus preventing explicit structural information from being diluted into vague semantic features.  HPDM successfully decouples the original input into a low-level geometric prior called $\mathrm{PCP}_1$ and a high-level semantic prior called $\mathrm{PCP}_2$, ensuring that only regions with isotropic structural consistency unique to the real target are amplified, while suppressing background clutter.

\subsubsection{Multi-directional Feature Aggregation Calculation Blocks (MFACBs)}
The objective is to quantify the centripetal convergence of infrared small targets by transforming raw intensity distributions into structured physical gradient fields. Unlike standard convolutional layers that learn arbitrary filters, MFACBs employ a physically constrained architecture to explicitly model the target's directional symmetry. For an input feature map $X \in \mathbb{R}^{B \times C \times H \times W}$, where $B$, $C$, $H$, and $W$ denote the batch size, channel number, height, and width, respectively, the block first extracts structural responses across discrete orientations $N_d = 24$ and
$\theta \in \{0^\circ, 15^\circ, \dots, 345^\circ\}$. In order to maintain physical interpretability and computational efficiency, we implement this as a depth-wise convolution (DWConv) with frozen weights,  and the directional gradient $F_{\theta}$ is defined as
\begin{equation}
    F_{\theta} = \mathrm{ReLU}( \mathrm{BN}( \mathrm{DWConv}(X, {K}_{\theta}) ) ), 
\end{equation}
where ${K}_\theta$ represents the pre-defined, non-trainable kernels. The operators $\mathrm{DWConv}$, $\mathrm{BN}$, and $\mathrm{ReLU}$ denote the depth-wise convolution, batch normalization, and ReLU activation, respectively. This frozen mechanism ensures that the backbone remains anchored to a consistent physical proxy, preventing structural primitives from being diluted by stochastic gradient updates during the early stages of training.

To integrate these discrete directional fields into a unified physical representation, a point-wise convolution (P-wise Conv) of depth $1 \times 1$  is utilized as a fusion layer. This layer learns the optimal linear combination of the 24 directional features. Then, these features are combined with the original input through a residual connection to produce the multi-directional gradient feature $F_\textrm{MD}$, which is given by
\begin{equation}
    F_\textrm{MD} = W_\textrm{fusion} \cdot \text{Concat}(F_{0^\circ}, F_{15^\circ}, \dots, F_{345^\circ}) + X,
\end{equation}
where $W_\textrm{fusion}$ denotes the learnable weights of the fusion layer, and \text{Concat} denotes channel-wise feature concatenation. Besides, HPDM deploys three MFACBs in parallel with distinct kernel scales $n \in \{3, 5, 7\}$. This multi-scale configuration ensures robust gradient perception for both sub-pixel point targets and larger structural objects. 

\begin{figure*}[t]
    \centering
    \includegraphics[width = 0.95\textwidth]{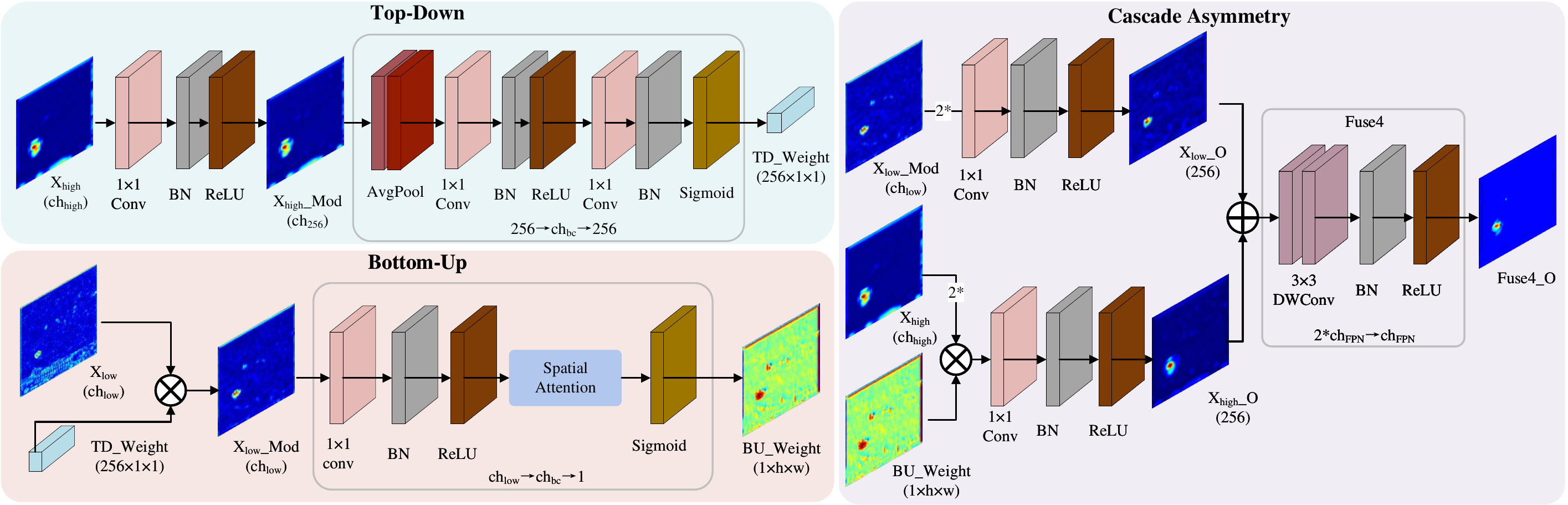} 
    \caption{Detailed structure of ISC-A, including the generation of top-down and bottom-up weights and the cascaded asymmetric interaction.} 
    \label{ISCA_structure} 
\end{figure*}

\subsubsection{Decoupled Priors Extraction Mechanism}
As shown in Fig. \ref{fig:HPDM}, the prior heads (PHs) are comprised of two distinct branches, which include $\mathrm{PH}_1$ for the spatial context prior and $\mathrm{PH}_2$ for the contour-aware prior head, which processes the concatenated multi-scale feature maps. To effectively facilitate hierarchical learning of targets, the concatenated outputs of MFACBs at three scales ($3 \times 3$, $5 \times 5$, and $7 \times 7$) are processed through dual-stream extraction modules to generate two complementary representations: an enhanced prior map $\mathrm{PCP}_1$ for precise target localization, and a set of multi-scale hierarchical features $\mathrm{PCP}_2$ for subsequent feature embedding. This process is mathematically represented as
\begin{equation}
    \begin{aligned}
        \mathrm{PCP}_1 &= \text{PH}_{1}( F_c( Fe (I_\mathrm{in}, 3), Fe(I_\mathrm{in}, 5), Fe(I_\mathrm{in}, 7) ) ), \\
        \mathrm{PCP}_{2i} &= \text{PH}_{2}( F_c( Fe(I_\mathrm{in}, 3), Fe(I_\mathrm{in}, 5), Fe(I_\mathrm{in}, 7) ) ),
    \end{aligned}
\end{equation}
where $\text{PH}_1(\cdot)$ and $\text{PH}_2(\cdot)$ represent the respective extraction streams, $F_c(\cdot)$ denotes the feature concatenation, $Fe(I_\mathrm{in},n)$ denotes the MFACB feature extraction at kernel scale $n\times n$, $I_\mathrm{in}$ is the input image, and $i$ indexes the hierarchical feature maps generated by $\text{PH}_2$.

To be more precise, $\mathrm{PH}_1$ serves as a fixed anchor-preserving projection branch that utilizes a $3 \times 3$ convolution and the $\mathrm{sigmoid}$ activation to produce a single-channel spatial significance map $\mathrm{PCP}_1$, providing stable target localization guidance unaffected by stochastic training noise. At the same time, $\mathrm{PH}_2$ employs a learnable architecture with max-pooling and sequential $3 \times 3$ convolutions to extract $\mathrm{PCP}_{2i}$ for subsequent feature embedding, a set of multi-scale contour-aware primitives that adaptively capture the target's topological energy distribution. By combining the stable localization of $\mathrm{PH}_1$ with the adaptive morphological perception of $\mathrm{PH}_2$, the dual-head design ensures that the backbone receives a comprehensive physical initialization that facilitates robust detection in complex cluttered environments.

\subsection{Dual Prior Interactive Fusion Module (DPIFM)}\label{DPIFM}

This module is designed to ensure that the extracted $\mathrm{PCP}_1$ and $\mathrm{PCP}_2$ can be seamlessly integrated into the optimal position of the features extracted by the backbone network. 
Specifically, $\mathrm{PCP}_1$ and $\mathrm{PCP}_2$ are up-sampled to match the spatial resolution of backbone features, then $\mathrm{PCPs}$ and the features are jointly input into the bidirectional prior guidance feature enhancer (BPGFE) to improve the target saliency. Since the backbone network generates multi-scale target features, multiple BPGFEs will output multi-scale feature maps to capture the different levels of target characteristics.

Fig. \ref{fig:BPGFE} shows the structure of our BPGFE. Initially, it performs a prior semantic calibration. Using the high-level context of the deep features, it generates a feature-driven attention map $M_F$ to spatially gate the raw prior maps $P \in \mathbb{R}^{9 \times H \times W}$ of $\mathrm{PCP}_1$ and $\mathrm{PCP}_2$. Here, $M_F$ and $P_\textrm{ref}$ are defined as
\begin{equation}
    \begin{aligned}
        & M_F = \sigma( W_{f2} \ast \delta( \mathcal{B}( W_{f1} \ast F ) ) ), \\
        & P_\textrm{ref} = P \otimes M_F,
    \end{aligned}
\end{equation}
where $\ast$ denotes the convolution, $W_{f1}$ and $W_{f2}$ denote the learnable $1 \times 1$ convolution-kernel parameters. Here, $\delta(\cdot)$ denotes the ReLU activation, $\mathcal{B}(\cdot)$ denotes the batch normalization, and $\otimes$ denotes the element-wise multiplication. This step suppresses background noise in the prior field that lacks semantic support, resulting in a refined prior $P_\textrm{ref}$ that is both physically grounded and semantically consistent.

Subsequently, BPGFE applies a prior to feature (F) extracted by backbone enhancement to reinforce target saliency within the deep feature space. It leverages the structural primitives preserved in $P_\textrm{ref}$ to produce a localized structural attention map $M_P$ as follows
\begin{equation}
    \begin{aligned}
        & M_P = \sigma ( W_{p2} \ast \delta ( \mathcal{B} ( W_{p1} \ast P_\textrm{ref}  )  )  ), \\
        & F_{\mathrm{weighted}} = F \otimes M_P.
    \end{aligned}
\end{equation}
Here, $W_{p1}$ and $W_{p2}$ denote the learnable convolution-kernel parameters, and the $3 \times 3$ convolution parameterized by $W_{p1}$ is employed in this branch. BPGFE effectively captures the local centripetal gradient geometry of the refined priors, using it to relocate the target's position in the feature map $F$. This prevents the erosion of small target signatures that typically occurs in standard deep networks.

Finally, BPGFE harmonizes these bidirectional flows by concatenating the weighted features and refined priors. A fusion convolution parameterized by $W_\textrm{fusion}$ is applied to aggregate these diverse information streams into an enhanced representation $F_\textrm{out}$, which is given by
\begin{equation}
    F_\textrm{out} = \delta ( \mathcal{B}( W_\textrm{fusion} \ast (F_\textrm{weighted}, P_\textrm{ref}) ) ).
\end{equation}
By deploying this bidirectional enhancement, our proposed PICANet perceives a persistent awareness of the target's physical attributes, allowing for precise detection and morphological reconstruction even in the presence of intense structural clutter.

\subsection{Multi-level Cross-Feature Attention Module (MCFAM)}\label{MCFAM}

The backbone network extracts multi-scale and multi-level target features for subsequent fusion to generate final prediction maps. However, up-sampling and pooling operations can lead to misalignment of semantic texture information and spatial location features. MCFAM employs a multi-level interactive spatial-channel attention (ISC-A) mechanism to align the detailed texture and spatial position information of each scale and level of the target, as shown in Fig. \ref{ISCA_structure}.

The ISC-A unit first performs a global-to-local calibration to identify channels that carry significant target responses. Given the encoder feature set $U = [u_1, \dots, u_C]$, a global descriptor extraction process $z_c$ is initiated to represent the global contextual energy as
\begin{equation}
    z_c = \text{Pool}_\text{global}(u_c) = \frac{1}{H \times W} \sum_{i=1}^{H} \sum_{j=1}^{W} u_c(i, j).
\end{equation}
Collecting these channel descriptors gives $z=[z_1,\ldots,z_C]^\top$. Then a channel calibration network (comprising two point-wise layers $W_1, W_2$ and a bottleneck ratio $r=4$) is used to generate the channel-wise attention vector $A_\text{ch}$ in the form of
\begin{equation}
    A_\text{ch} = \sigma(\text{BN}(W_2 \cdot \text{ReLU}(\text{BN}(W_1 \cdot z)))).
\end{equation}
By applying this vector to the encoder features $F^l$, the unit produces a semantically filtered feature $F_\mathrm{SC-A} = A_\text{ch} \otimes F^l$, effectively muting background channels that lack target-like characteristics.
To ensure a precise reconstruction of minuscule targets, the ISC-A unit further executes spatial context distillation. This stage utilizes refined gradient information to guide the decoder’s feature representation. Local peaks and regional statistics from $F_\mathrm{SC-A}$ are aggregated using a dual pooling strategy. Let $F_{\mathrm{filtered}}=F_\mathrm{SC-A}$ denote the channel-filtered feature supplied to this spatial distillation stage. Thus, the result is
\begin{equation}
    \begin{aligned}
        M_\text{spatial} = \delta\left(\mathrm{Conv}_{3 \times 3}\left( 
        \begin{aligned}
            \mathrm{AvgPool}(F_{\mathrm{filtered}}) \\
            \mathrm{MaxPool}(F_{\mathrm{filtered}})
        \end{aligned} \right)\right)  
    \end{aligned}. 
\end{equation}
The final interactive spatial cross feature $F_\mathrm{ISC-A}$ is obtained by modulating the high-level features of the decoder $F^h$ with this distilled spatial map, and $F_\mathrm{ISC-A}$ is given by
\begin{equation}
    F_\mathrm{ISC-A} = M_\text{spatial} \otimes F^h.
\end{equation}

Note that our  PICANet adopts a collaborative prediction strategy to integrate information from different stages. By aggregating the ISC-A representations across multiple network levels, MCFAM ensures that the physical priors, which are preserved and refined by HPDM and DPIFM, remain effective throughout the decoding process right up to the final pixel prediction. The collaborative multi-scale output is given by
\begin{equation}
    O = \sum_{k=1}^{K} \mathrm{Up}^k( \Phi( F_\mathrm{SC-A}^k, F_\mathrm{ISC-A}^k ) ),
\end{equation}
where $K$ denotes the number of prediction levels, $k$ indexes the prediction level, $\mathrm{Up}^k$ denotes the up-sampling operation to the original resolution, and $\Phi(\cdot)$ represents the fusion of the dual-path interactive features.

\subsection{Loss Function}\label{Loss Function}

The training process is supervised by a composite loss function which is designed to balance global background suppression, local structural fidelity, and pixel-level segmentation precision. Specifically, the loss function is formulated as 
\begin{equation}
    \begin{aligned}
        L_{\mathrm{total}} & = \omega_\text{bce}(\alpha) \cdot L_{\mathrm{mBCE}} + \omega_\text{iou}(\alpha) \cdot L_{\mathrm{SoftIoU}} \\ & \quad + \omega_\text{prior} \cdot L_{\mathrm{MaskedMSE}}.
    \end{aligned}
\end{equation}
Here, $L_{\mathrm{mBCE}}$ is the multi-scale binary cross-entropy loss, i.e.,
\begin{equation}
    L_{\mathrm{mBCE}} = \sum_{n=1}^{6} \mathrm{BCE}(\sigma (O^n), Y),
\end{equation}
where $\sigma$ denotes the sigmoid function, $n$ indexes the prediction level, and $Y$ is the ground-truth mask. $L_{\mathrm{SoftIoU}}$ is defined as
\begin{equation}
L_{\mathrm{SoftIoU}} = 1 - \frac{1}{N_b} \sum_{b=1}^{N_b} \mathrm{SIoU}_b, 
\end{equation}
where $N_b$ denotes the batch size, $\mathrm{SIoU}_b$ represents the soft intersection-over-union for sample $b$ \cite{rahman2016optimizing,wu2024rpcanet}.
$L_{\mathrm{MaskedMSE}}$ is the masked mean squared error loss, which is defined as
\begin{equation}
    \begin{aligned}
        &L_{\mathrm{MaskedMSE}} \\
        &= \sum_{i,j} \| \Phi_\text{feat}(i,j) - \Phi_\text{prior}(i,j) \|^2 \cdot \frac{M(i,j)}{\displaystyle\sum_{i,j} M(i,j) + \epsilon},
    \end{aligned}
\end{equation}
where $\Phi_\text{feat}$ denotes the deep features, $\Phi_\text{prior}$ ($\mathrm{PCP}_2$) denotes the physical priors, $M$ is a target-neighborhood mask, $i$ and $j$ index spatial rows and columns, and $\epsilon$ is a small positive constant for numerical stability.

In addition, we introduce a dynamic weighting strategy that moves from coarse localization to fine-grained refinement. The weights $\omega_\text{bce}(\alpha)$ and $\omega_\text{iou}(\alpha)$ are determined according to
\begin{equation}
    \begin{cases}
        \omega_\text{bce}(\alpha) = 15.0 \times (1.0 - 0.6 \alpha), \\
        \omega_\text{iou}(\alpha) = 10.0 + 40.0 \alpha,
    \end{cases}
\end{equation}
where $\alpha = t / T$ represents the training ratio, with $t$ and $T$ denoting the current epoch and the total epochs, respectively. Furthermore, $\omega_\text{prior}$ is fixed at 1.0 and is not scaled by $\alpha$.

\begin{table*}[t]
    \centering
    \setlength{\tabcolsep}{4pt}
    \renewcommand{\arraystretch}{1.2}
    \caption{Performance comparisons of different methods in terms of $\mathrm{mIoU}$, $\mathrm{F_1}$, $\mathrm{P_d}$, and $\mathrm{F_a}$ on NUDT-SIRST, IRSTD-1k, and SIRST-Aug, where the best and second-best results are highlighted in \textcolor{red}{red} and \textcolor{blue}{blue}, respectively. }
    \label{performance_comparison}
    \resizebox{\textwidth}{!}{
    \begin{tabular}{l cccc cccc cccc cc c}
        \hline 
        \multirow{2}{*}{Method}
        & \multicolumn{4}{c}{NUDT-SIRST} 
        & \multicolumn{4}{c}{IRSTD-1k} 
        & \multicolumn{4}{c}{SIRST-Aug}
        & Params & FLOPs & Time (s) \\
        \cmidrule(lr){2-5} \cmidrule(lr){6-9} \cmidrule(lr){10-13} 
        & $\mathrm{mIoU}$ $\uparrow$ & $\mathrm{F_1}$ $\uparrow$ & $\mathrm{P_d}$ $\uparrow$ & $\mathrm{F_a}$ $\downarrow$ 
        & $\mathrm{mIoU}$ $\uparrow$ & $\mathrm{F_1}$ $\uparrow$ & $\mathrm{P_d}$ $\uparrow$ & $\mathrm{F_a}$ $\downarrow$ 
        & $\mathrm{mIoU}$ $\uparrow$ & $\mathrm{F_1}$ $\uparrow$ & $\mathrm{P_d}$ $\uparrow$ & $\mathrm{F_a}$ $\downarrow$ 
        & (M) & (G) & CPU/GPU \\
        \hline  \hline
        Top-hat \cite{bai2010analysis} & 22.36 & 36.38 & 91.38 & 771.20 & 5.59 & 10.67 & 69.54 & 102.30 & 16.39 & 28.62 & 81.26 & 166.30 & -- & -- & 0.0018/-- \\
        IPI \cite{gao2013infrared} & 34.83  & 51.49 & 92.58 & 106.90 & 18.67 & 31.48 & 78.54 & 93.37 & 21.90 & 35.97 & 80.36 & 93.17 & -- & -- & 3.0972/-- \\
        MPCM \cite{wei2016multiscale} & 25.96 & 40.78 & 78.59 & 316.50 & 14.81 & 25.93 & 69.03 & 64.24 & 19.49 & 33.00 & 93.58 & 304.80 & -- & -- & 0.0624/-- \\
        PSTNN \cite{zhang2019infrared} & 25.46 & 40.58 & 78.52 & 36.50 & 14.87 & 25.89 & 68.73 & 83.51 & 19.76 & 33.00 & 93.40 & 51.96 & -- & -- & 0.3490/-- \\
        \hline
        ACM \cite{dai2021asymmetric} & 70.02 & 82.54 & 95.79 & 16.89 & 62.97 & 77.37 & 90.75 & 17.83 & 70.87 & 82.36 & 95.84 & 43.09 & 0.40 & 0.40 & --/0.0072 \\
        ALCNet \cite{dai2021attentional} & 85.03 & 91.91 & 97.89 & 18.20 & 63.52 & 77.61 & 85.86 & \textcolor{red}{12.96} & 69.49 & 81.73 & 95.32 & 69.32 & 0.38 & 3.74 & --/0.0103 \\
        ISNet \cite{zhang2022isnet} & 87.77 & 93.37 & 95.56 & 9.42 & 64.73 & 77.99 & 87.21 & \textcolor{blue}{15.03} & 70.12 & 82.42 & 92.30 & \textcolor{red}{21.57} & 0.97 & 30.62 & --/0.0173 \\
        AGPCNet \cite{zhang2023attention} & 85.31 & 92.45 & 97.90 & 4.77 & 61.00 & 75.75 & 89.35 & 25.34 & 72.36 & 83.83 & 99.03 & 35.56 & 12.36 & 43.18 & --/0.0205 \\
        UIUNet \cite{wu2023uiu} & 88.71 & 94.01 & 91.43 & 8.91 & 63.06 & 77.35 & 90.60 & 20.57 & 71.80 & 83.59 & 98.35 & 28.29 & 50.54 & 54.42 & --/0.0317 \\
        RDIAN \cite{sun2023receptive} & 85.71 & 92.01 & 97.43 & 22.91 & 66.06 & 79.35 & 91.60 & 22.57 & 71.97 & 84.02 & 98.65 & 54.32 & 0.02 & 3.72 & --/0.0248 \\
        SCTransNet \cite{yuan2024sctransnet} & 91.06 & 93.15 & 95.43 & 11.19 & 64.38 & 79.55 & 92.18 & 27.46 & 72.98 & 82.99 & 91.84 & 38.75 & 5.74   & 20.24 & --/0.0213 \\
        $\mathrm{L}^{2}\mathrm{SKNet}$ \cite{wu2024saliency} & 93.26 & 96.67 & 97.69 & 5.19 & 67.81 & 80.81 & 90.24 & 27.46 & 73.97 & 85.01 & 99.23 & 55.10 & 1.07  & 6.01 & --/0.0122 \\
        MSHNet \cite{liu2024infrared} & 89.99 & 93.57 & 96.07 & 9.63 & 64.50 & 77.55 & 91.68 & 34.46 & 71.64 & 84.16 & 90.78 & \textcolor{blue}{23.09} & 4.07  & 6.11 & --/0.0245 \\ 
        DURD \cite{hao2026deep} & 93.85 & 96.83 & 98.19 & 5.56  & 66.79 & 80.46 & 92.94 & 30.76 & 73.19 & 84.84 & 98.89 & 33.41 & 2.46 & 52.84 & --/0.0206 \\ \hline
        RPCANet \cite{wu2024rpcanet} & 89.31 & 94.35 & 97.14 & 28.72  & 63.21 & 77.45 & 88.31 & 43.90 & 72.54 & 84.08 & 98.21 & 34.14 & 0.68  & 27.48 & --/0.0096 \\
        DRPCANet \cite{xiong2025drpca} & 93.12 & 96.02 & 98.02 & 19.50  & 63.93 & 78.15 & 92.09 & 49.20 & 73.93 & 85.39 & 98.12 & 30.45 & 1.17  & 91.05 & --/0.0101 \\
        CTVNet \cite{liu2025ctvnet} & 92.05 & 94.91 & 98.16 & 8.97  & 66.35 & 77.16 & 92.84 & 20.47 & 72.96 & 85.16 & 99.04 & 38.79 & 14.90 & 85.74 & --/0.0195 \\
        RPCANet++ \cite{wu2025rpcanet++} & 92.46 & 96.05 & 98.05 & 14.45  & 64.03 & 77.26 & 89.35 & 42.80 & 73.14 & 84.39 & 97.36 & 32.48 & 1.44  & 31.21 & --/0.0063 \\
        L-RPCANet \cite{liu2025lightweight} & 94.39 & 97.11 & \textcolor{blue}{99.15} & 1.39 & 67.44 & 80.05 & 93.94 & 43.90 & 75.35 & 85.93 & \textcolor{blue}{99.35} & 26.43 & 0.22 & 8.59 & --/0.0029 \\
        \hline
        \rowcolor{gray!10} PICANet (U) & \textcolor{red}{95.19} & \textcolor{red}{97.32} & \textcolor{red}{99.24} & \textcolor{red}{1.28} & \textcolor{red}{70.31} & \textcolor{red}{82.96} &\textcolor{red}{94.62} & 23.38 & \textcolor{blue}{75.48} & \textcolor{blue}{86.39} & \textcolor{red}{99.39} & 28.56 & 6.33 & 8.53 & --/0.0129 \\
        \rowcolor{gray!10} PICANet (F) & \textcolor{blue}{94.69} & \textcolor{blue}{97.27} & 98.31 & \textcolor{blue}{1.34} & \textcolor{blue}{69.35} & \textcolor{blue}{81.93} & \textcolor{blue}{94.05} & 29.43 & \textcolor{red}{75.58} & \textcolor{red}{86.44} & 98.62 & 25.95 & 2.74 & 4.27 & --/0.0089 \\
        \hline
    \end{tabular}
    }
\end{table*}
\subsection{Plug-and-Play}\label{Implementation on FPN and U-Net}

As previously mentioned, our proposed PICANet offers the advantage of plug-and-play capability. To demonstrate this property, we integrate the above modules into U-Net \cite{ronneberger2015u} and the ResNet-based feature pyramid network (ResNet-FPN) \cite{lin2017feature}. In the U-Net configuration, the network relies on a symmetric encoder-decoder structure for semantic segmentation. Standard skip connections transfer raw background noise directly to the decoder. To resolve this, MCFAM units are deployed across all skip paths. These units actively filter the low-level encoder features using high-level semantic cues from the decoder. Then, DPIFM is embedded in each encoder stage to continuously fuse the decoupled physical priors with the down-sampled features. For the ResNet-FPN setting, ResNet-50 serves as the primary backbone to extract multi-scale feature pyramids. The integration strategy here adapts to the top-down pathway design. HPDM is positioned at the input stage to extract the initial geometric and semantic priors from the raw image. Rather than using conventional lateral connections with simple element-wise addition, FPN integrates MCFAM units. These units strictly align the top-down up-sampled features with their corresponding lateral backbone features. Meanwhile, DPIFM modulates the multi-scale representations at each residual stage. 

Beyond these structural changes, the framework replaces conventional single-head outputs with a collaborative prediction strategy. As illustrated in Fig. \ref{prior_extract_fuse_flow}, the model produces a local target estimate $O^k$ at each decoding stage. The final output $O_\text{final}$ aggregates these multi-scale target contexts. This collaborative mechanism allows the model to combine the high-level semantic confidence of deep layers (e.g., $O^1, O^2$) with the precise spatial localization of shallower layers (e.g., $O^5, O^6$). The aggregation is defined as
\begin{equation}
    O_\text{final} = \mathcal{S} ( \sum_{k=1}^{K} \mathrm{Up}^k (O^k) ),
\end{equation}
where  $\mathcal{S}$ is the fusion operator producing the final pixel-level confidence map.

\section{Experiments}\label{Experiment}

In this section, our proposed PICANet is integrated with U-Net and ResNet-FPN, denoted as PICANet (U) and PICANet (F), respectively. Numerical experiments are conducted to compare them with benchmark \textit{model-driven} methods including Top-hat \cite{bai2010analysis}, IPI \cite{gao2013infrared}, MPCM \cite{wei2016multiscale}, PSTNN \cite{zhang2019infrared}, \textit{data-driven} methods including ACM \cite{dai2021asymmetric}, ALCNet \cite{dai2021attentional}, ISNet \cite{zhang2022isnet}, AGPCNet \cite{zhang2023attention}, UIUNet \cite{wu2023uiu}, RDIAN \cite{sun2023receptive}, SCTransNet \cite{yuan2024sctransnet}, $\mathrm{L}^{2}\mathrm{SKNet}$ \cite{wu2024saliency}, MSHNet \cite{liu2024infrared}, DURD \cite{hao2026deep}, and \textit{model-data-driven} methods including RPCANet \cite{wu2024rpcanet}, DRPCANet \cite{xiong2025drpca}, CTVNet \cite{liu2025ctvnet}, RPCANet++ \cite{wu2025rpcanet++}, L-RPCANet \cite{liu2025lightweight}.

Subsection \ref{Experimental} introduces the experimental setups. Subsection \ref{Numerical} gives the numerical results. Subsection \ref{Ablation} presents the ablation experiments. Subsection \ref{Discussion} provides some discussion.

\subsection{Experimental Setups}\label{Experimental}

\subsubsection{Dataset Description}

All evaluations are carried out on three public datasets NUDT-SIRST\footnote{\url{https://github.com/YeRen123455/Infrared-Small-Target-Detection}}, IRSTD-1k\footnote{\url{https://github.com/RuiZhang97/ISNet}}, and SIRST-Aug\footnote{\url{https://github.com/Tianfang-Zhang/AGPCNet}}.
These datasets encompass a variety of real-world and synthetic infrared imaging scenes, exhibiting significant differences in target size, intensity, and shape, while also reflecting diverse background complexities and sensor characteristics. Additionally, the image resolution is 256 $\times$ 256 pixels for the NUDT-SIRST and SIRST-Aug datasets, whereas it is 512 $\times$ 512 pixels for the IRSTD-1k dataset.

\begin{figure*}[t]
    \centering
    \includegraphics[width=0.96\textwidth]{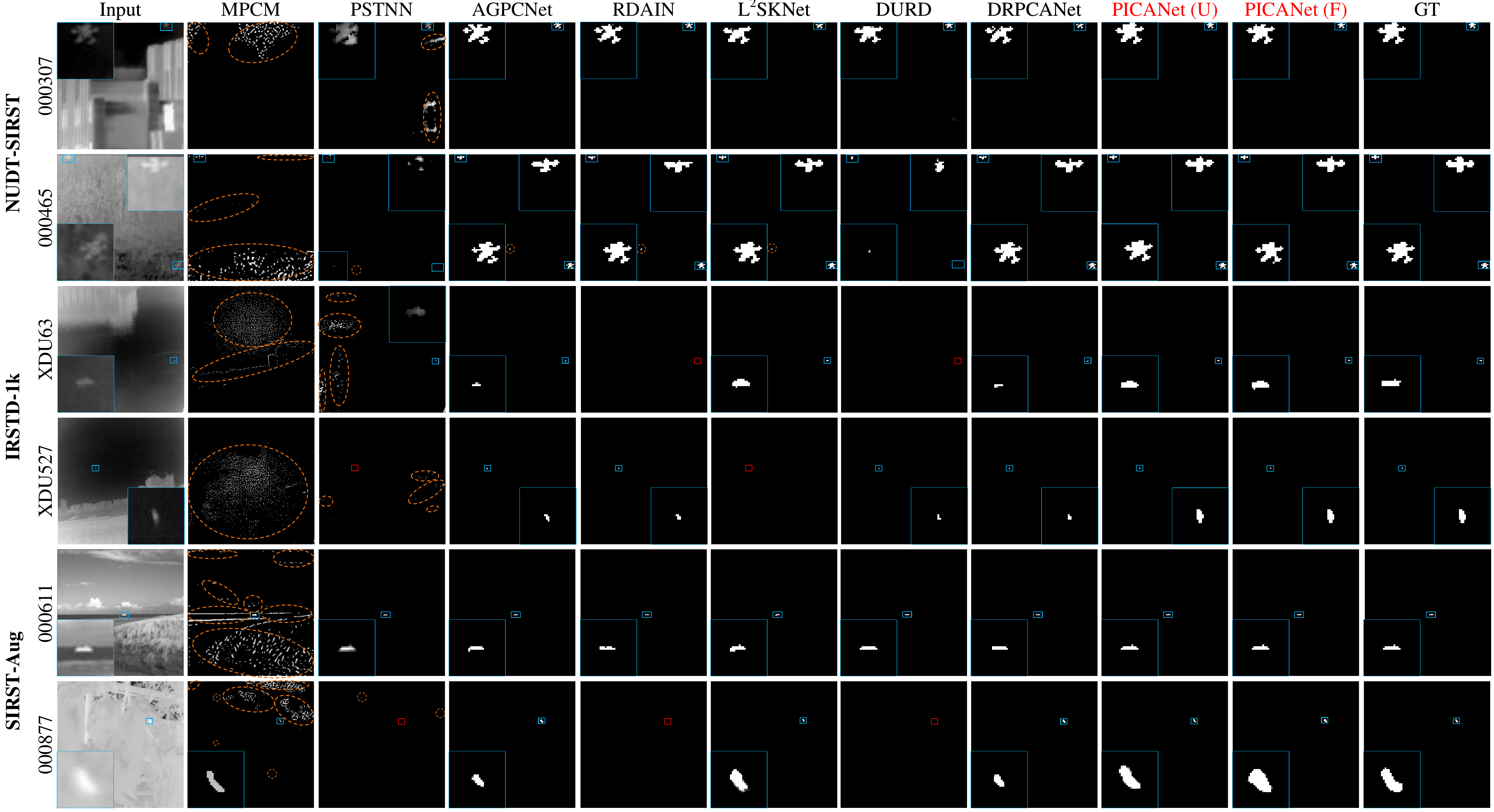}
    \caption{Representative visual results on the three datasets. The targets are represented by rectangles colored in blue, orange, and red, indicating true positive targets, false positive targets, and false negative targets, respectively.}
\label{fig:Representative_visual_comparison}
\end{figure*}


\subsubsection{Implementation Details} 
All experiments are conducted on a workstation equipped with an Intel Xeon Gold 6526Y CPU and an Nvidia GeForce 4090 GPU. Our PICANet and the deep learning methods are retrained for 400 epochs with a batch size of 8 in the PyTorch framework. The compared methods strictly follow the default learning rates and configurations from their official codes. Our PICANet utilizes the Adam optimizer with an initial learning rate of $1 \times 10^{-4}$, which automatically decreases via a polynomial decay scheduler. 

To ensure a fair evaluation, all compared methods share the same training and testing set splits. Data augmentation operations include random horizontal flipping, vertical flipping, and transposition, each applied with a probability of 0.5. This augmentation strategy is applied  exclusively to the SIRST-Aug dataset, while the other datasets undergo no augmentation. The input resolutions for the NUDT-SIRST, IRSTD-1k, and SIRST-Aug datasets align with their default dimensions, which also determine the resolution of the final visualization results. Furthermore, floating‑point operations (FLOPs) are calculated  based on a random input tensor of size $1 \times 1 \times 256 \times 256$. Inference time is measured on the CPU for \textit{model-driven} methods and the GPU for \textit{data-driven} methods and \textit{model-data-driven} methods. This metric is evaluated by averaging the forward pass duration per image over the testing set, excluding data loading after a standard warm-up phase.

\subsubsection{Evaluation Metrics}
As suggested in \cite{wu2024rpcanet}, four metrics are considered to evaluate target segmentation and infrared small target detection.

\begin{itemize}
    \item Mean intersection over union (mIoU) is a pixel-level evaluation for semantic segmentation. Let $M$ and $\mathrm{IoU}_c$ be the total number of categories and the intersection over union on category $c$. Then $\mathrm{mIoU}$ is defined as
\begin{equation}\label{Eq.(18)}
\mathrm{mIoU} = \frac{1}{M} \sum_{c=1}^{M} \mathrm{IoU}_c.
\end{equation}
    \item $\mathrm{F_1}$-score ($\mathrm{F_1}$) is the harmonic mean of precision and recall. Let $\mathrm{TP}$, $\mathrm{FP}$, and $\mathrm{FN}$ denote true-positive, false-positive, and false-negative pixels, respectively. $\mathrm{F_1}$ is defined as
 \begin{equation}    \label{Eq.(19)}
    \mathrm{F_1} = \frac{2\, \mathrm{TP}}{2\mathrm{TP}+\mathrm{FP}+\mathrm{FN}}.
 \end{equation}
    \item Probability of detection ($\mathrm{P_d}$) evaluates the proportion of targets correctly detected by the target detection model among all real targets. $\mathrm{P_d}$ is defined as
\begin{equation}
    \mathrm{P_d} = \frac{\mathrm{TP}}{\mathrm{TP} + \mathrm{FN}}.
    \label{Eq.(20)}
\end{equation}

    \item False alarm rate ($\mathrm{F_a}$) measures the ratio of falsely predicted pixels ($\mathrm{P}_{\mathrm{false}}$) to all image pixels ($\mathrm{P}_{\mathrm{all}}$), which is defined as
\begin{equation}
    \mathrm{F_a} = \frac {\mathrm{P}_{\mathrm{false}}}{\mathrm{P}_{\mathrm{all}}}.
    \label{Eq.(21)}
\end{equation}
\end{itemize}

Therefore, $\mathrm{mIoU}$ is adopted to evaluate the performance of the multi-category image segmentation model, $\mathrm{F_1}$ to evaluate the balance between precision and recall, $\mathrm{P_d}$, and $\mathrm{F_a}$ to evaluate localization ability.

\begin{table*}[t]
    \centering
    \renewcommand{\arraystretch}{1.2} 
    \caption{Ablation experiments based on PCPs and BPGFE in terms of $\mathrm{mIoU}$, $\mathrm{F_1}$, $\mathrm{P_d}$, and $\mathrm{F_a}$ on NUDT-SIRST, IRSTD-1k, and SIRST-Aug, where the best and second-best results are highlighted in \textcolor{red}{red} and \textcolor{blue}{blue}, respectively.}
    \label{tab:ablation_dpifm}
    \resizebox{\linewidth}{!}{
    \begin{tabular}{l ccc cccc cccc cccc}
        \hline
        \multirow{2}{*}{Backbone} & \multirow{2}{*}{$\mathrm{PCP}_1$} & \multirow{2}{*}{$\mathrm{PCP}_2$} & \multirow{2}{*}{BPGFE}
        & \multicolumn{4}{c}{NUDT-SIRST} & \multicolumn{4}{c}{IRSTD-1k} & \multicolumn{4}{c}{SIRST-Aug} \\
        \cmidrule(lr){5-8} \cmidrule(lr){9-12} \cmidrule(lr){13-16}
        & & & & $\mathrm{mIoU}\uparrow$ & $\mathrm{F_1}\uparrow$ & $\mathrm{P_d}\uparrow$ & $\mathrm{F_a}\downarrow$ 
        & $\mathrm{mIoU}\uparrow$ & $\mathrm{F_1}\uparrow$ & $\mathrm{P_d}\uparrow$ & $\mathrm{F_a}\downarrow$ 
        & $\mathrm{mIoU}\uparrow$ & $\mathrm{F_1}\uparrow$ & $\mathrm{P_d}\uparrow$ & $\mathrm{F_a}\downarrow$ \\
        \hline\hline
        \multirow{5}{*}{U-Net}
        & \ding{55} & \ding{55} & \ding{55} & 86.32 & 90.56 & 88.32 & 18.65 & 59.12 & 65.56 & 86.44 & 60.38 & 63.41 & 73.01 & 87.45 & 83.36 \\
        & \ding{51} & \ding{55} & \ding{55} & 89.29 & 92.68 & 94.99 & 8.41 & 63.66 & 70.34 & 89.77 & 48.56 & 66.83 & 77.56 & 92.53 & 48.89 \\
        & \ding{55} & \ding{51} & \ding{55} & 89.52 & 92.89 & 95.63 & 9.52 & 64.11 & 71.25 & 90.27 & 50.19 & 67.22 & 77.91 & 93.06 & 50.08 \\
        & \ding{51} & \ding{51} & \ding{55} & 90.78 & 93.33 & 93.78 & 10.56 & 64.78 & 73.56 & 91.76 & 40.81 & 69.54 & 79.49 & 94.56 & 40.65 \\
        \rowcolor{gray!10}
        & \ding{51} & \ding{51} & \ding{51} 
        & \textcolor{red}{95.19} & \textcolor{red}{97.32} & \textcolor{red}{99.24} & \textcolor{red}{1.28} & \textcolor{red}{70.31} & \textcolor{red}{82.96} &\textcolor{red}{94.62} & \textcolor{red}{23.38} & \textcolor{blue}{75.48} & \textcolor{blue}{86.39} & \textcolor{red}{99.39} & \textcolor{blue}{28.56} \\
        \hline\noalign{\vskip 2pt}\hline %
        \multirow{5}{*}{ResNet-FPN}
        & \ding{55} & \ding{55} & \ding{55} & 85.16 & 89.35 & 87.89 & 28.33 & 58.52 & 64.51 & 86.52 & 62.45 & 61.78 & 71.79 & 86.07 & 81.97 \\
        & \ding{51} & \ding{55} & \ding{55} & 87.57 & 90.62 & 93.03 & 10.69 & 61.93 & 69.56 & 88.21 & 49.62 & 65.45 & 77.02 & 91.16 & 68.62 \\
        & \ding{55} & \ding{51} & \ding{55} & 88.23 & 90.97 & 93.66 & 10.25 & 62.37 & 70.27 & 89.33 & 48.12 & 66.19 & 77.65 & 91.86 & 64.20 \\
        & \ding{51} & \ding{51} & \ding{55} & 89.46 & 92.94 & 94.07 & 6.83 & 65.32 & 74.94 & 92.32 & 35.33 & 68.93 & 80.68 & 94.62 & 53.55 \\
        \rowcolor{gray!10}
        & \ding{51} & \ding{51} & \ding{51} 
        &  \textcolor{blue}{94.69} & \textcolor{blue}{97.27} & \textcolor{blue}{98.31} & \textcolor{blue}{1.34} & \textcolor{blue}{69.35} & \textcolor{blue}{81.93} & \textcolor{blue}{94.05} & \textcolor{blue}{29.43} & \textcolor{red}{75.58} & \textcolor{red}{86.44} & \textcolor{blue}{98.62} & \textcolor{red}{25.95} \\
        \hline
    \end{tabular}
    }
\end{table*}

\begin{figure*}[t]
    \centering
    \includegraphics[width=0.95\textwidth]{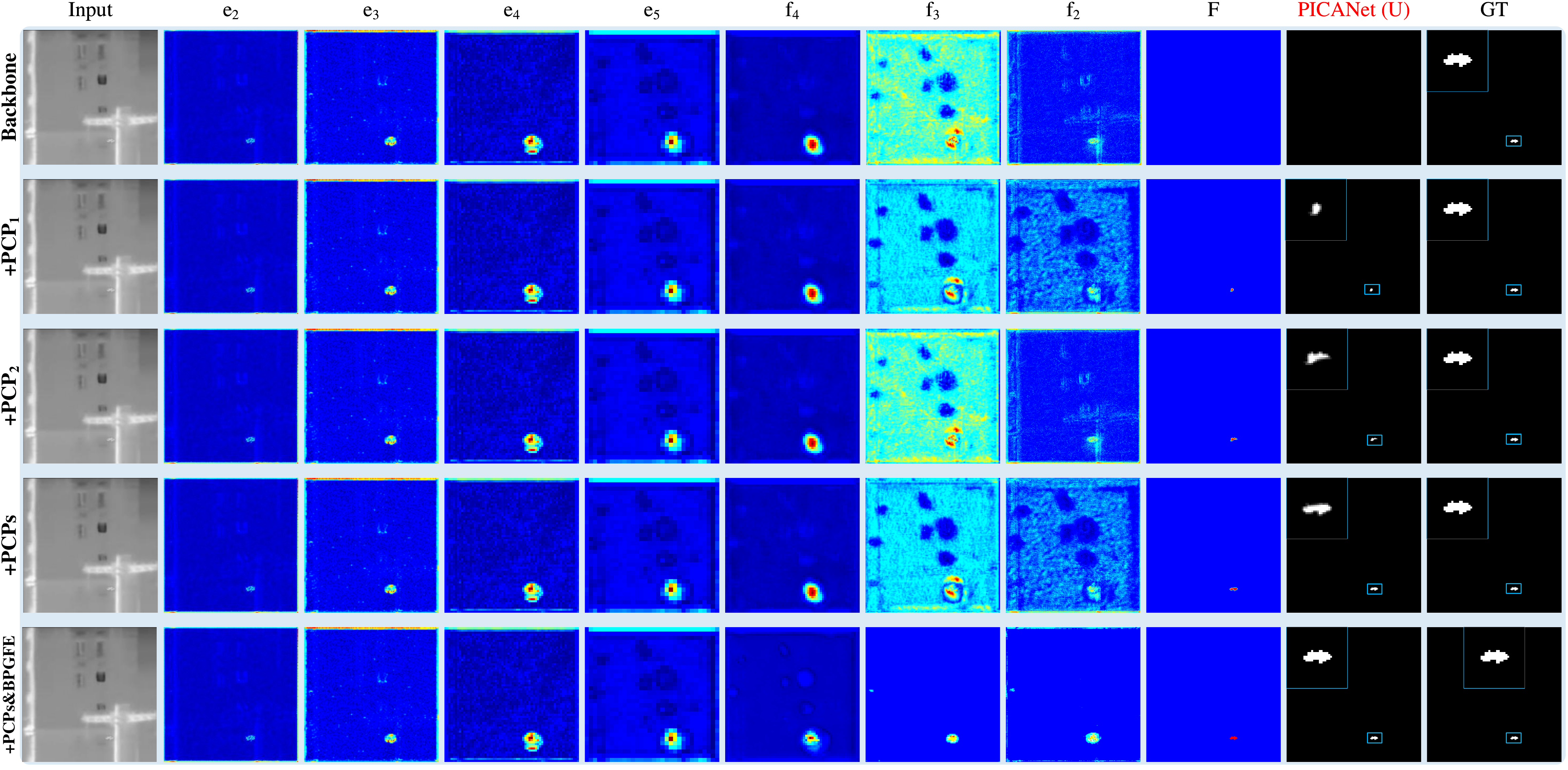}
    \caption{Ablation studies on the effectiveness of PCPs and BPGFE in the U-Net-based model. The first to fifth rows represent backbone, backbone + $\mathrm{PCP}_1$, backbone + $\mathrm{PCP}_2$, backbone + PCPs, and backbone + PCPs \& BPGFE, respectively. }
    \label{fig:ablation_DPIFM_UNet}
\end{figure*}

\begin{figure*}[t]
    \centering
    \includegraphics[width=0.95\textwidth]{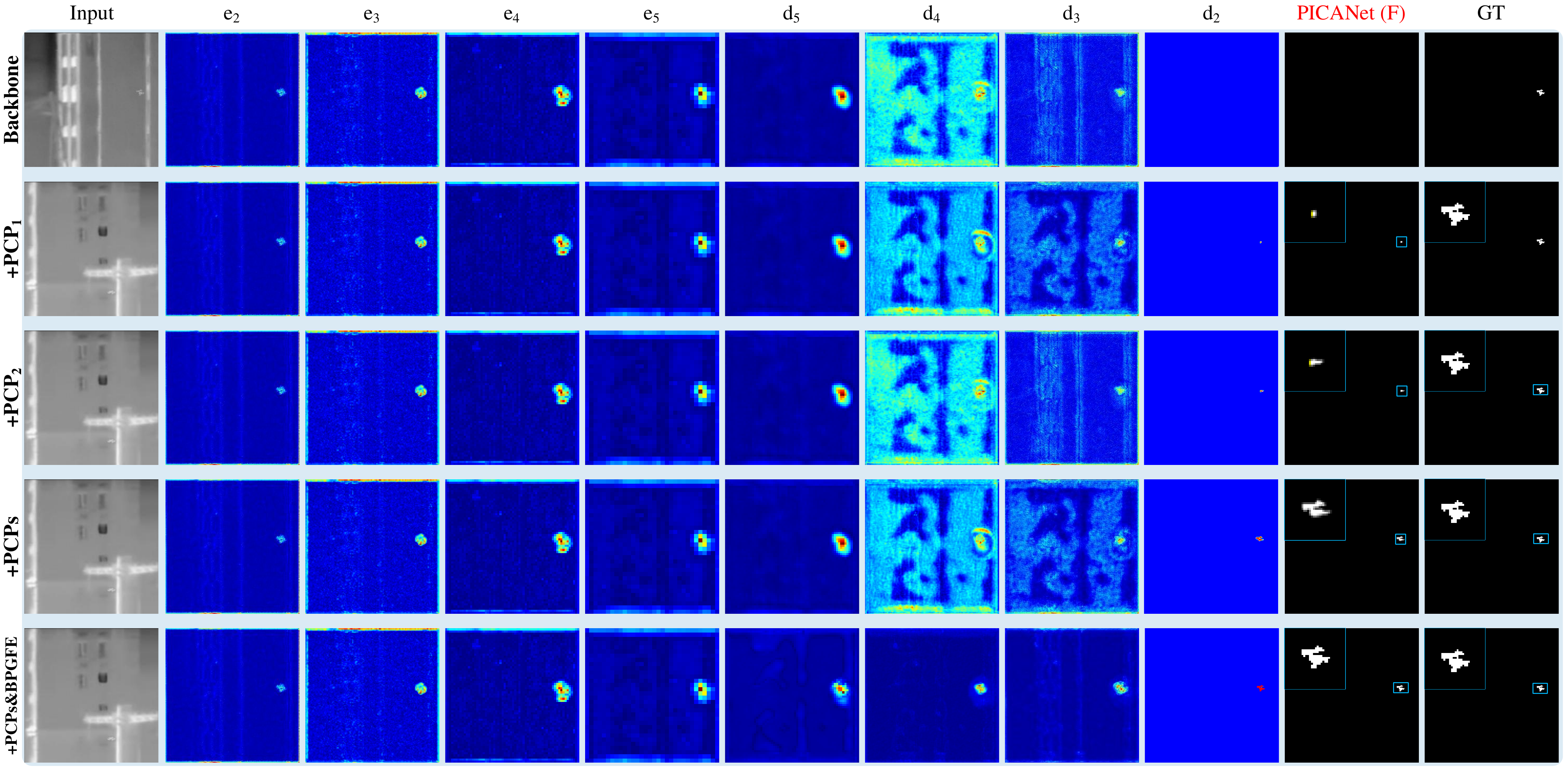}
    \caption{Ablation studies on the effectiveness of PCPs and BPGFE in the ResNet-FPN-based model. The first to fifth rows represent backbone, backbone + $\mathrm{PCP}_1$, backbone + $\mathrm{PCP}_2$, backbone + PCPs, and backbone + PCPs \& BPGFE, respectively. }
    \label{fig:ablation_DPIFM_FPN}
\end{figure*}

\subsection{Numerical Results} \label{Numerical}

\subsubsection{Quantitative Comparison}

Table \ref{performance_comparison} lists the quantitative results of all compared methods, where the best and second-best results are highlighted in \textcolor{red}{red} and \textcolor{blue}{blue}, respectively. Note that the symbols $\uparrow$ and $\downarrow$ indicate whether a higher or lower value is preferable.

Generally speaking, \textit{model-driven} methods rely on fixed physical priors and struggle to accurately characterize complex contours against complex backgrounds, resulting in a noticeable performance gap. Although \textit{data-driven} methods such as UIUNet can achieve higher detection rates, they often suffer from overfitting and domain instability due to the lack of physical guidance. \textit{Model-data-driven} methods such as RPCANet and L-RPCANet combine the strengths of both approaches and demonstrate impressive detection performance. However, they often mechanically assign equal weights to all unfolding stages, failing to distinguish the relative importance of background features versus target features. 

In contrast, our proposed PICANet (U) and PICANet (F) illustrate excellent performance across three public datasets, achieving a favorable balance among Params, FLOPs, and Time. Specifically, on the NUDT-SIRST dataset, PICANet (U) ranks first across all metrics, while PICANet (F) remains among the top-performing methods. In the challenging IRSTD-1k dataset, PICANet (U) and PICANet (F) also sit in the top positions except for $\mathrm{F_a}$. Note that they have substantial improvement in $\mathrm{mIoU}$, reaching 70.31\% and 69.35\%, respectively. For the SIRST-Aug dataset, PICANet (F) demonstrates superior semantic extraction with the best $\mathrm{mIoU}$ and $\mathrm{F_1}$, while PICANet (U) maintains the highest $\mathrm{P_d}$. These results clearly indicate that physical priors are crucial for the ISTD task.

\begin{table*}[t]
    \centering
    \renewcommand{\arraystretch}{1.2} 
    \caption{Ablation experiments of ISC-A in terms of $\mathrm{mIoU}$, $\mathrm{F_1}$, $\mathrm{P_d}$, and $\mathrm{F_a}$ on NUDT-SIRST, IRSTD-1k, and SIRST-Aug, where the best and second-best results are highlighted in \textcolor{red}{red} and \textcolor{blue}{blue}, respectively.}
    \label{tab:ablation_ISC_A}
    \begin{tabular}{l ccc cccc cccc cccc}
        \hline
        \multirow{2}{*}{Backbone} & \multirow{2}{*}{TD} & \multirow{2}{*}{BU} & \multirow{2}{*}{CA}
        & \multicolumn{4}{c}{NUDT-SIRST} & \multicolumn{4}{c}{IRSTD-1k} & \multicolumn{4}{c}{SIRST-Aug} \\
        \cmidrule(lr){5-8} \cmidrule(lr){9-12} \cmidrule(lr){13-16}
        & & & & $\mathrm{mIoU}\uparrow$ & $\mathrm{F_1}\uparrow$ & $\mathrm{P_d}\uparrow$ & $\mathrm{F_a}\downarrow$ 
        & $\mathrm{mIoU}\uparrow$ & $\mathrm{F_1}\uparrow$ & $\mathrm{P_d}\uparrow$ & $\mathrm{F_a}\downarrow$ 
        & $\mathrm{mIoU}\uparrow$ & $\mathrm{F_1}\uparrow$ & $\mathrm{P_d}\uparrow$ & $\mathrm{F_a}\downarrow$ \\
        \hline\hline
        \multirow{5}{*}{U-Net}
        & \ding{55} & \ding{55} & \ding{55} & 88.13 & 92.31 & 90.07 & 17.01 & 60.74 & 67.32 & 88.10 & 58.72 & 65.09 & 74.74 & 89.11 & 81.69 \\
        & \ding{51} & \ding{55} & \ding{55} & 91.16 & 94.42 & 96.75 & 7.04  & 65.43 & 72.11 & 91.45 & 46.88 & 68.62 & 79.31 & 94.22 & 47.21 \\
        & \ding{55} & \ding{51} & \ding{55} & 91.45 & 94.68 & 97.40 & 6.83  & 65.94 & 73.07 & 91.98 & 46.03 & 69.07 & 79.73 & 94.78 & 46.13 \\
        & \ding{51} & \ding{51} & \ding{55} & 92.74 & 95.19 & 95.54 & 5.92  & 66.68 & 75.45 & 93.51 & 38.94 & 71.47 & 81.37 & 96.32 & 38.96 \\
        \rowcolor{gray!10}
        & \ding{51} & \ding{51} & \ding{51} 
        & \textcolor{red}{95.19} & \textcolor{red}{97.32} & \textcolor{red}{99.24} & \textcolor{red}{1.28} & \textcolor{red}{70.31} & \textcolor{red}{82.96} &\textcolor{red}{94.62} & \textcolor{red}{23.38} & \textcolor{blue}{75.48} & \textcolor{blue}{86.39} & \textcolor{red}{99.39} & \textcolor{blue}{28.56} \\
        \hline\noalign{\vskip 2pt}\hline %
        \multirow{5}{*}{ResNet-FPN}
        & \ding{55} & \ding{55} & \ding{55} & 87.42 & 91.68 & 90.27 & 26.05 & 60.84 & 66.83 & 88.80 & 60.14 & 64.13 & 74.17 & 88.45 & 79.61 \\
        & \ding{51} & \ding{55} & \ding{55} & 89.97 & 93.06 & 95.44 & 8.36  & 64.39 & 71.98 & 90.63 & 47.23 & 67.94 & 79.49 & 93.58 & 66.23 \\
        & \ding{55} & \ding{51} & \ding{55} & 90.68 & 93.45 & 96.12 & 7.84  & 64.87 & 72.75 & 91.80 & 45.69 & 68.73 & 80.17 & 94.34 & 61.76 \\
        & \ding{51} & \ding{51} & \ding{55} & 91.96 & 95.41 & 96.59 & 4.38  & 67.76 & 77.39 & 94.80 & 32.86 & 71.57 & 83.20 & 97.15 & 51.02 \\
        \rowcolor{gray!10}
        & \ding{51} & \ding{51} & \ding{51} 
        &  \textcolor{blue}{94.69} & \textcolor{blue}{97.27} & \textcolor{blue}{98.31} & \textcolor{blue}{1.34} & \textcolor{blue}{69.35} & \textcolor{blue}{81.93} & \textcolor{blue}{94.05} & \textcolor{blue}{29.43} & \textcolor{red}{75.58} & \textcolor{red}{86.44} & \textcolor{blue}{98.62} & \textcolor{red}{25.95} \\
        \hline
    \end{tabular}
\end{table*}

\begin{table*}[t]
    \centering
    \caption{Discussion on different prior knowledge embeddings in terms of $\mathrm{mIoU}$, $\mathrm{F_1}$, $\mathrm{P_d}$, and $\mathrm{F_a}$ on NUDT-SIRST, IRSTD-1k, and SIRST-Aug, where the best and second-best results are highlighted in \textcolor{red}{red} and \textcolor{blue}{blue}, respectively.}
    \label{Ablation_experiments_on_the_HPDM}
     \renewcommand{\arraystretch}{1.2}
    \begin{tabular}{l l cccc cccc cccc}
    \hline
    \multirow{2}{*}{Backbone} & \multirow{2}{*}{Method}
    & \multicolumn{4}{c}{NUDT-SIRST}
    & \multicolumn{4}{c}{IRSTD-1k}
    & \multicolumn{4}{c}{SIRST-Aug} \\
    \cmidrule(lr){3-6} \cmidrule(lr){7-10} \cmidrule(lr){11-14}
    &
    & $\mathrm{mIoU}\uparrow$ & $\mathrm{F_1}\uparrow$ & $\mathrm{P_d}\uparrow$ & $\mathrm{F_a}\downarrow$
    & $\mathrm{mIoU}\uparrow$ & $\mathrm{F_1}\uparrow$ & $\mathrm{P_d}\uparrow$ & $\mathrm{F_a}\downarrow$
    & $\mathrm{mIoU}\uparrow$ & $\mathrm{F_1}\uparrow$ & $\mathrm{P_d}\uparrow$ & $\mathrm{F_a}\downarrow$ \\
    \hline\hline
    \multirow{5}{*}{U-Net}
    & Top-hat
    & 86.65 & 92.82 & 96.14 & 33.91
    & 65.58 & 79.21 & 91.24 & 44.40
    & 70.12 & 78.94 & 89.89 & 68.93 \\
    & IPI
    & 83.81 & 91.24 & 96.51 & 63.84
    & 64.57 & 78.47 & 89.23 & 54.81
    & 70.45 & 79.99 & 90.46 & 67.88 \\
    & MPCM
    & 89.23 & 94.30 & 97.14 & 30.91
    & 64.85 & 78.68 & 94.28 & 46.02
    & 69.83 & 78.65 & 90.63 & 79.36 \\
    & PSTNN
    & 82.13 & 90.21 & 96.00 & 33.83
    & 63.53 & 77.69 & 91.25 & 46.23
    & 68.36 & 79.45 & 89.56 & 80.52 \\
    \rowcolor{gray!10}
    & HPDM
    & \textcolor{red}{95.19} & \textcolor{red}{97.32} & \textcolor{red}{99.24} & \textcolor{red}{1.28} & \textcolor{red}{70.31} & \textcolor{red}{82.96} &\textcolor{red}{94.62} & \textcolor{red}{23.38} & \textcolor{blue}{75.48} & \textcolor{blue}{86.39} & \textcolor{red}{99.39} & \textcolor{blue}{28.56} \\
    \hline\noalign{\vskip 2pt}\hline
    \multirow{5}{*}{ResNet-FPN}
    & Top-hat
    & 86.65 & 92.82 & 96.14 & 33.91
    & 65.58 & 79.21 & 91.24 & 44.40
    & 70.12 & 78.94 & 89.89 & 68.93 \\
    & IPI
    & 83.81 & 91.24 & 96.51 & 63.84
    & 64.57 & 78.47 & 89.23 & 54.81
    & 70.45 & 79.99 & 90.46 & 67.88 \\
    & MPCM
    & 89.23 & 94.30 & 97.14 & 30.91
    & 64.85 & 78.68 & 94.28 & 46.02
    & 69.83 & 78.65 & 90.63 & 79.36 \\
    & PSTNN
    & 82.13 & 90.21 & 96.00 & 33.83
    & 63.53 & 77.69 & 91.25 & 46.23
    & 68.36 & 79.45 & 89.56 & 80.52 \\
    \rowcolor{gray!10}
    & HPDM
    & \textcolor{blue}{94.69} & \textcolor{blue}{97.27} & \textcolor{blue}{98.31} & \textcolor{blue}{1.34} & \textcolor{blue}{69.35} & \textcolor{blue}{81.93} & \textcolor{blue}{94.05} & \textcolor{blue}{29.43} & \textcolor{red}{75.58} & \textcolor{red}{86.44} & \textcolor{blue}{98.62} & \textcolor{red}{25.95} \\
    \hline
    \end{tabular}
\end{table*}

\subsubsection{Visual Comparison}

To further demonstrate the superiority of our proposed PICANet, a visual comparison is conducted across these three datasets, as shown in Fig. \ref{fig:Representative_visual_comparison}. Here, we select \textit{model-driven} methods (MPCM, PSTNN), \textit{data-driven} methods (AGPCNet, RDIAN, $\mathrm{L}^{2}\mathrm{SKNet}$, DURD), and the \textit{model-data-driven} method (DRPCANet).

It can be seen that \textit{model-driven} methods exhibit significant limitations when confronted with non-uniform backgrounds. For instance, MPCM is highly susceptible to high-brightness noise and edges, resulting in an overwhelming number of false positives (orange rectangles). Although \textit{data-driven} and \textit{model-data-driven} methods  achieve superior localization performance, they struggle to preserve target morphological integrity when dealing with targets located near image edges or extremely faint signals, often leading to shape erosion or fragmentation. In contrast, PICANet consistently delivers robust performance across all scenarios by effectively leveraging decoupled physical priors. Driven by $\mathrm{PCP}_1$ and DPIFM's active gating mechanism, the framework accurately locks onto spatial coordinates (blue boxes) near image boundaries or under low signal-to-clutter ratio conditions, strictly filtering out background clutter. Simultaneously, guided by $\mathrm{PCP}_2$ and MCFAM's collaborative prediction strategy, it achieves exceptional morphological fidelity, reconstructing target shapes that precisely match the ground truth. This visually confirms that our proposed physics-informed paradigm successfully bridges deterministic physical laws with modern deep learning, establishing a highly reliable solution for high-fidelity ISTD.

\subsection{Ablation Studies} \label{Ablation}


\subsubsection{Effect of DPIFM}

Table \ref{tab:ablation_dpifm} presents the quantitative results in both U-Net and ResNet-FPN backbones. Regardless of the chosen backbone architecture, a consistent performance trend is observed. When $\mathrm{PCP}_1$ and $\mathrm{PCP}_2$ are introduced into the network without  DPIFM, the performance improvements remain marginal. This indicates that passively adding physical priors to deep networks without an adaptive gating mechanism fails to fully exploit their potential. However, when DPIFM is enabled to actively modulate these collaborative priors, the detection accuracy experiences a significant leap. With the complete configuration, both U-Net and ResNet-FPN variants achieve optimal values across all datasets. This confirms that DPIFM perfectly complements the physical knowledge extracted by HPDM, acting as a crucial bridge for high-fidelity target reconstruction.

Fig. \ref{fig:ablation_DPIFM_UNet} and Fig. \ref{fig:ablation_DPIFM_FPN} illustrate the feature evolution under different ablation settings for the U-Net and ResNet-FPN architectures. From left to right, the columns present the input image, encoder features $e_2-e_5$, decoder features $f_4-f_2$ or $d_4-d_1$, prediction, and ground truth. The heatmap colors range from blue to red, representing low to high energy. The encoder features remain similar across all settings, indicating that the encoding process captures target semantics and background clutter together. The backbone alone cannot suppress complex structural interference. Adding $\mathrm{PCP}_1$ or $\mathrm{PCP}_2$ partially reduces background clutter, but noise artifacts remain in the decoder. Their direct combination in PCPs is still insufficient to remove high-intensity false alarms. In contrast, the introduction of $\mathrm{PCPs} \& \mathrm{BPGFE}$ leverages  prior-driven dynamic gating to progressively suppress background clutter during decoding, producing focused target energy and predictions that closely match the ground truth.

\subsubsection{Effect of ISC-A}
The proposed ISC-A module performs bidirectional feature purification through a cascaded asymmetric routing mechanism. To validate the effectiveness, we conduct rigorous ablation experiments, as reported in Table \ref{tab:ablation_ISC_A}. It can be observed that enabling top-down channel modulation  or bottom-up spatial modulation  independently improves performance for both U-Net and ResNet-FPN, confirming the benefits of channel noise filtration and spatial target enrichment. The parallel symmetric configuration in the fourth row provides limited gains because TD and BU operate independently without interaction. With cascade asymmetry   enabled in the fifth row, the spatial mask is derived after channel-wise background suppression, allowing the full ISC-A module to achieve the highest $\mathrm{mIoU}$ and $\mathrm{F_1}$ while substantially reducing the false alarm rate $\mathrm{F_a}$. The same trend holds for both backbones, demonstrating that ISC-A is a robust, plug-and-play module with strong architecture generalization.

\subsection{Discussion} \label{Discussion}

In fact, HPDM aims to extract two collaborative physical priors, i.e., $\mathrm{PCP}_1$ and $\mathrm{PCP}_2$, for morphological fidelity. 
This discussion is conducted on U-Net and ResNet-FPN, and the results are provided in Table \ref{Ablation_experiments_on_the_HPDM}.

It is evident that the HPDM framework shows overwhelming superiority over previous prior embeddings. When integrated into the U-Net backbone, the HPDM setting achieves optimal values on the NUDT-SIRST and IRSTD-1k datasets, dominating key metrics such as $\mathrm{mIoU}$, $\mathrm{F_1}$, and $\mathrm{P_d}$, while suppressing $\mathrm{F_a}$ to an absolute minimum. Furthermore, this performance advantage remains completely consistent across different architectures. Upon transitioning to the ResNet-FPN backbone, the HPDM variant continues to secure the best overall metrics, particularly excelling on the SIRST-Aug dataset. In contrast, traditional physical models provide rigid and limited prior information, leading to degraded performance regardless of the chosen backbone. These results show that HPDM extracts highly adaptive physical constraints, and its effectiveness is completely independent of the underlying network topology.

\section{Conclusion}\label{Conclusion}

In this paper, we propose PICANet, a physics-informed cascaded asymmetric network that seamlessly bridges traditional physical laws with deep representation learning for ISTD. Unlike existing methods, PICANet is a plug-and-play framework that effectively resolves the misalignment of multi-level features in different backbone networks. Its consistent performance across U-Net and ResNet-FPN demonstrates strong compatibility and generalization across various backbones. Extensive evaluations on public datasets demonstrate that the proposed PICANet achieves comparable or superior detection performance to benchmark \textit{model-driven}, \textit{data-driven}, and \textit{model-data-driven} methods.

Although our PICANet provides plug-and-play compatibility across different backbone networks, it also incurs computational overhead. Our future work will focus on developing a more lightweight plug-and-play framework that maintains effective multi-level feature alignment and detection performance while improving computational efficiency.

\bibliographystyle{IEEEtran}
\bibliography{mybibfile}

@article{liu2025enhancing,
  title={Enhancing infrared small target detection: A saliency-guided multi-task learning approach},
  author={Liu, Zhaoying and Zhang, Yuxiang and He, Junran and Zhang, Ting and ur Rehman, Sadaqat and Saraee, Mohamad and Sun, Changming},
  journal={IEEE Transactions on Intelligent Transportation Systems},
  volume={26},
  number={3},
  pages={3603--3618},
  year={2025},
  publisher={IEEE}
}

@article{liu2025lightweight,
  title={Lightweight deep unfolding networks with enhanced robustness for infrared small target detection},
  author={Liu, Jingjing and Han, Yinchao and Xiu, Xianchao and Zhang, Jianhua and Liu, Wanquan},
  journal={arXiv preprint arXiv:2509.08205},
  year={2025}
}

@inproceedings{ronneberger2015u,
  title={U-net: Convolutional networks for biomedical image segmentation},
  author={Ronneberger, Olaf and Fischer, Philipp and Brox, Thomas},
  booktitle={International Conference on Medical Image Computing and Computer-Assisted Intervention},
  pages={234--241},
  year={2015},
  organization={Springer}
}

@article{li2024usesnet,
  author={Li, Lingxiao and Liu, Linlin and He, Yunan and Zhong, Zhuqiang},
  title={{USES-Net}: An Infrared Dim and Small Target Detection Network with Embedded Knowledge Priors},
  journal={Electronics}, volume={13}, number={7}, pages={1400}, year={2024},
  doi={10.3390/electronics13071400}
}

@inproceedings{lin2017feature,
  title={Feature pyramid networks for object detection},
  author={Lin, Tsung-Yi and Doll{\'a}r, Piotr and Girshick, Ross and He, Kaiming and Hariharan, Bharath and Belongie, Serge},
  booktitle={Proceedings of the IEEE conference on Computer Vision and Pattern Recognition},
  pages={2117--2125},
  year={2017}
}

@article{deng2025cspenet,
  title={{CSPEN}et: Contour-Aware and Saliency Priors Embedding Network for Infrared Small Target Detection},
  author={Deng, Jiakun and Li, Kexuan and Cui, Xingye and Li, Jiaxuan and Long, Chang and Pu, Tian and Peng, Zhenming},
  journal={IEEE Transactions on Geoscience and Remote Sensing},
  year={2025},
  volume={63},
  number={},
  pages={1--14},
  publisher={IEEE}
}

@article{hua2025survey,
  title={A survey of small object detection based on deep learning in aerial images},
  author={Hua, Wei and Chen, Qili},
  journal={Artificial Intelligence Review},
  volume={58},
  number={6},
  pages={162},
  year={2025},
  publisher={Springer}
}

@article{yuan2024sctransnet,
  title={Sc{T}rans{N}et: Spatial-channel cross transformer network for infrared small target detection},
  author={Yuan, Shuai and Qin, Hanlin and Yan, Xiang and Akhtar, Naveed and Mian, Ajmal},
  journal={IEEE Transactions on Geoscience and Remote Sensing},
  volume={62},
  pages={1--15},
  year={2024},
  publisher={IEEE}
}

@article{kumar2025small,
  title={Small and dim target detection in infrared imagery: A review, current techniques and future directions},
  author={Kumar, Nikhil and Singh, Pravendra},
  journal={Neurocomputing},
  pages={129640},
  year={2025},
  publisher={Elsevier}
}

@article{li2025ilnet,
  title={{ILNet}: Low-level matters for salient infrared small target detection},
  author={Li, Haoqing and Yang, Jinfu and Wang, Runshi and Xu, Yifei},
  journal={IEEE Transactions on Aerospace and Electronic Systems},
  volume={61},
  number={4},
  pages={8306--8318},
  year={2025},
  publisher={IEEE}
}

@article{liu2025graph,
  title={Graph {L}aplacian regularization for fast infrared small target detection},
  author={Liu, Ting and Liu, Yongxian and Yang, Jungang and Li, Boyang and Wang, Yingqian and An, Wei},
  journal={Pattern Recognition},
  volume={158},
  pages={111077},
  year={2025},
  publisher={Elsevier}
}

@article{zhang2021agpcnet,
  title={{AGPCN}et: Attention-guided pyramid context networks for detecting infrared small target under complex background},
  author={Zhang, Tianfang and Li, Lei and Cao, Siying and Pu, Tian and Peng, Zhenming},
  journal={IEEE Transactions on Aerospace and Electronic Systems},
  volume={59},
  number={4},
  pages={4250--4261},
  year={2023},
  publisher={IEEE}
}

@inproceedings{zhang2022isnet,
  title={{ISNet}: Shape matters for infrared small target detection},
  author={Zhang, Mingjin and Zhang, Rui and Yang, Yuxiang and Bai, Haichen and Zhang, Jing and Guo, Jie},
  booktitle={Proceedings of the IEEE/CVF Conference on Computer Vision and Pattern Recognition},
  pages={877--886},
  year={2022}
}

@article{sui2026samamba,
  title={{SAMamba}: Stream Alignment {M}amba for Motion Infrared Small Target Detection},
  author={Sui, Yi and Zhi, Xiyang and Huang, Yuanxin and Shi, Tianjun and Jiang, Shikai},
  journal={IEEE Transactions on Circuits and Systems for Video Technology},
   year={2026},
  volume={36},
  number={8},
  pages={11215-11230},
  publisher={IEEE}
}

@article{lu2026hyperistd,
  title={{HyperISTD}: Angular-Consistent Modeling in Hyperbolic Space for Infrared Small Target Detection},
  author={Lu, Yubing and Liu, Pingping and Zhang, Tongshun and Li, Aohua and Zhou, Qiuzhan},
  journal={IEEE Transactions on Circuits and Systems for Video Technology},
  year={2026},
  publisher={IEEE}
}

@article{gao2013infrared,
  title={Infrared patch-image model for small target detection in a single image},
  author={Gao, Chenqiang and Meng, Deyu and Yang, Yi and Wang, Yongtao and Zhou, Xiaofang and Hauptmann, Alexander G},
  journal={IEEE Transactions on Image Processing}, 
  volume={22},
  number={12},
  pages={4996--5009},
  year={2013},
  publisher={IEEE}
}

@article{wei2016multiscale,
  title={Multiscale patch-based contrast measure for small infrared target detection},
  author={Wei, Yantao and You, Xinge and Li, Hong},
  journal={Pattern Recognition},
  volume={58},
  pages={216--226},
  year={2016},
  publisher={Elsevier}
}

@article{zhang2019infrared,
  title={Infrared small target detection based on partial sum of the tensor nuclear norm},
  author={Zhang, Landan and Peng, Zhenming},
  journal={Remote Sensing},
  volume={11},
  number={4},
  pages={382},
  year={2019},
  publisher={MDPI}
}

@article{wu2023uiu,
  title={{UIU}-{N}et: {U}-{N}et in {U}-{N}et for infrared small object detection},
  author={Wu, Xin and Hong, Danfeng and Chanussot, Jocelyn},
  journal={IEEE Transactions on Image Processing},
  volume={32},
  pages={364--376},
  year={2023},
  publisher={IEEE}
}

@article{sun2023receptive,
  title={Receptive-field and direction induced attention network for infrared dim small target detection with a large-scale dataset IRDST},
  author={Sun, Heng and Bai, Junxiang and Yang, Fan and Bai, Xiangzhi},
  journal={IEEE Transactions on Geoscience and Remote Sensing},
  volume={61},
  pages={1--13},
  year={2023},
  publisher={IEEE}
}

@article{chen2022local,
  title={Local patch network with global attention for infrared small target detection},
  author={Chen, Fang and Gao, Chenqiang and Liu, Fangcen and Zhao, Yue and Zhou, Yuxi and Meng, Deyu and Zuo, Wangmeng},
  journal={IEEE Transactions on Aerospace and Electronic Systems},
  volume={58},
  number={5},
  pages={3979--3991},
  year={2022},
  publisher={IEEE}
}

@inproceedings{liu2024infrared,
  title={Infrared Small Target Detection with Scale and Location Sensitivity},
  author={Liu, Qiankun and Liu, Rui and Zheng, Bolun and Wang, Hongkui and Fu, Ying},
  booktitle={Proceedings of the IEEE/CVF Computer Vision and Pattern Recognition},
  year={2024}
}

@article{zhang2023attention,
  title={Attention-guided pyramid context networks for detecting infrared small target under complex background},
  author={Zhang, Tianfang and Li, Lei and Cao, Siying and Pu, Tian and Peng, Zhenming},
  journal={IEEE Transactions on Aerospace and Electronic Systems},
  volume={59},
  number={4},
  pages={4250--4261},
  year={2023},
  publisher={IEEE}
}

@article{xiong2025drpca,
  title={{DRPCA}-{N}et: Make Robust {PCA} Great Again for Infrared Small Target Detection},
  author={Xiong, Zihao and Zhou, Fei and Wu, Fengyi and Yuan, Shuai and Fu, Maixia and Peng, Zhenming and Yang, Jian and Dai, Yimian},
  journal={IEEE Transactions on Geoscience and Remote Sensing},
  volume={63},
  number={ },
  pages={1--16},
  year={2025},
  publisher={IEEE}
}

@inproceedings{ren2021dnanet,
  title={{DNAN}et: Dense nested attention network for single image dehazing},
  author={Ren, Dongdong and Li, Jinbao and Han, Meng and Shu, Minglei},
  booktitle={ICASSP 2021-2021 IEEE International Conference on Acoustics, Speech and Signal Processing (ICASSP)},
  pages={2035--2039},
  year={2021},
  organization={IEEE}
}

@article{tong2021eaau,
  title={{EAAU-N}et: Enhanced asymmetric attention U-Net for infrared small target detection},
  author={Tong, Xiaozhong and Sun, Bei and Wei, Junyu and Zuo, Zhen and Su, Shaojing},
  journal={Remote Sensing},
  volume={13},
  number={16},
  pages={3200},
  year={2021},
  publisher={MDPI}
}

@article{liu2025ctvnet,
  title={{CTVN}et: Gradient Prior-Guided Deep Unfolding Network for Infrared Small Target Detection},
  author={Liu, Pei and Pang, Li and Peng, Jiangjun and Luo, Yisi and Liu, Junmin and Cao, Xiangyong},
  journal={IEEE Transactions on Geoscience and Remote Sensing},
  volume={63},
  pages={1--14},
  year={2025},
  publisher={IEEE}
}

@article{wu2026rpcanet,
  title={{RPCANet++}: Deep Interpretable Robust {PCA} for Sparse Object Segmentation},
  author={Wu, Fengyi and Dai, Yimian and Zhang, Tianfang and Ding, Yixuan and Yang, Jian and Cheng, Ming-Ming and Peng, Zhenming},
  journal={IEEE Transactions on Pattern Analysis and Machine Intelligence},
  year={2026},
  publisher={IEEE}
}

@article{qin2025osformer,
  title={{OSF}ormer: One-step transformer for infrared video small object detection},
  author={Qin, Haolin and Xu, Tingfa and Tang, Yuan and Xu, Fengxiang and Li, Jianan},
  journal={IEEE Transactions on Image Processing},
  year={2025},
  volume={34},
  number={},
  pages={5725--5736},
  publisher={IEEE}
}

@article{liu2023infrared,
  title={Infrared small and dim target detection with transformer under complex backgrounds},
  author={Liu, Fangcen and Gao, Chenqiang and Chen, Fang and Meng, Deyu and Zuo, Wangmeng and Gao, Xinbo},
  journal={IEEE Transactions on Image Processing},
  volume={32},
  pages={5921--5932},
  year={2023},
  publisher={IEEE}
}

@article{jiang2026bidirectional,
  title={Bidirectional Motion Perception {T}ransformer Network for Multi-frame Infrared Small Target Detection},
  author={Jiang, Zhiwei and Wang, Xiumei and Li, Xiangyang and Tian, Chunna and Zhou, Heng and Zhang, Zhenxi and Chen, Shiguo},
  journal={IEEE Transactions on Circuits and Systems for Video Technology},
  year={2026},
  volume={36},
  number={9},
  pages={12903--12916},
  publisher={IEEE}
}

@inproceedings{yuan2026seeing,
  title={Seeing through the noise: Improving infrared small target detection and segmentation from noise suppression perspective},
  author={Yuan, Maoxun and Meng, Duanni and Xi, Ziteng and Zhao, Tianyi and Zhao, Shiji and Dai, Yimian and Wei, Xingxing},
  booktitle={Proceedings of the IEEE/CVF Conference on Computer Vision and Pattern Recognition},
  pages={27783--27792},
  year={2026}
}

@inproceedings{wu2024rpcanet,
  title={{RPCANet}: Deep unfolding {RPCA} based infrared small target detection},
  author={Wu, Fengyi and Zhang, Tianfang and Li, Lei and Huang, Yian and Peng, Zhenming},
  booktitle={Proceedings of the IEEE/CVF Winter Conference on Applications of Computer Vision},
  pages={4809--4818},
  year={2024}
}

@article{wu2025rpcanet++,
  title={{RPCANet}++: Deep Interpretable Robust {PCA} for Sparse Object Segmentation},
  author={Wu, Fengyi and Dai, Yimian and Zhang, Tianfang and Ding, Yixuan and Yang, Jian and Cheng, Ming-Ming and Peng, Zhenming},
  journal={IEEE Transactions on Pattern Analysis and Machine Intelligence},
  year={2026},
  publisher={IEEE}
}

@article{luo2024revisiting,
  title={Revisiting nonlocal self-similarity from continuous representation},
  author={Luo, Yisi and Zhao, Xile and Meng, Deyu},
  journal={IEEE Transactions on Pattern Analysis and Machine Intelligence},
  volume={47},
  number={1},
  pages={450--468},
  year={2025},
  publisher={IEEE}
}

@inproceedings{rahman2016optimizing,
  title={Optimizing intersection-over-union in deep neural networks for image segmentation},
  author={Rahman, Md Atiqur and Wang, Yang},
  booktitle={International Symposium on Visual Computing},
  pages={234--244},
  year={2016},
  organization={Springer}
}

@article{yang2025deep,
  title={Deep learning based infrared small object segmentation: Challenges and future directions},
  author={Yang, Zhengeng and Yu, Hongshan and Zhang, Jianjun and Tang, Qiang and Mian, Ajmal},
  journal={Information Fusion},
  volume={118},
  pages={103007},
  year={2025},
  publisher={Elsevier}
}

@inproceedings{zhang2024irprunedet,
  title={{IRP}rune{D}et: efficient infrared small target detection via wavelet structure-regularized soft channel pruning},
  author={Zhang, Mingjin and Yang, Handi and Guo, Jie and Li, Yunsong and Gao, Xinbo and Zhang, Jing},
  booktitle={Proceedings of the AAAI Conference on Artificial Intelligence},
  volume={38},
  number={7},
  pages={7224--7232},
  year={2024}
}

@article{yu2022infrared,
  title={Infrared small target detection based on multiscale local contrast learning networks},
  author={Yu, Chuang and Liu, Yunpeng and Wu, Shuhang and Hu, Zhuhua and Xia, Xin and Lan, Deyan and Liu, Xin},
  journal={Infrared Physics \& Technology},
  volume={123},
  pages={104107},
  year={2022},
  publisher={Elsevier}
}

@article{bai2010infrared,
  title={Infrared small target enhancement and detection based on modified top-hat transformations},
  author={Bai, Xiangzhi and Zhou, Fugen},
  journal={Computers \& Electrical Engineering},
  volume={36},
  number={6},
  pages={1193--1201},
  year={2010},
  publisher={Elsevier}
}

@article{zhao2020infrared,
  title={Infrared small-target detection based on multiple morphological profiles},
  author={Zhao, Mingjing and Li, Lu and Li, Wei and Tao, Ran and Li, Liwei and Zhang, Wenjuan},
  journal={IEEE Transactions on Geoscience and Remote Sensing},
  volume={59},
  number={7},
  pages={6077--6091},
  year={2020},
  publisher={IEEE}
}

@article{bai2018derivative,
  title={Derivative entropy-based contrast measure for infrared small-target detection},
  author={Bai, Xiangzhi and Bi, Yanguang},
  journal={IEEE Transactions on Geoscience and Remote Sensing},
  volume={56},
  number={4},
  pages={2452--2466},
  year={2018},
  publisher={IEEE}
}

@article{wang2025paying,
  title={Paying more attention to local contrast: Improving infrared small target detection performance via prior knowledge},
  author={Wang, Peichao and Wang, Jiabao and Chen, Yao and Zhang, Rui and Li, Yang and Miao, Zhuang},
  journal={Engineering Applications of Artificial Intelligence},
  volume={146},
  pages={110244},
  year={2025},
  publisher={Elsevier}
}

@article{dai2021attentional,
  title={Attentional local contrast networks for infrared small target detection},
  author={Dai, Yimian and Wu, Yiquan and Zhou, Fei and Barnard, Kobus},
  journal={IEEE Transactions on Geoscience and Remote Sensing},
  volume={59},
  number={11},
  pages={9813--9824},
  year={2021},
  publisher={IEEE}
}

@article{wu2024saliency,
  title={Saliency at the helm: Steering infrared small target detection with learnable kernels},
  author={Wu, Fengyi and Liu, Anran and Zhang, Tianfang and Zhang, Luping and Luo, Junhai and Peng, Zhenming},
  journal={IEEE Transactions on Geoscience and Remote Sensing},
  volume={63},
  pages={1--14},
  year={2024},
  publisher={IEEE}
}

@article{kong2021infrared,
  title={Infrared small target detection via nonconvex tensor fibered rank approximation},
  author={Kong, Xuan and Yang, Chunping and Cao, Siying and Li, Chaohai and Peng, Zhenming},
  journal={IEEE Transactions on Geoscience and Remote Sensing},
  volume={60},
  pages={1--21},
  year={2021},
  publisher={IEEE}
}

@inproceedings{dai2021asymmetric,
  title={Asymmetric contextual modulation for infrared small target detection},
  author={Dai, Yimian and Wu, Yiquan and Zhou, Fei and Barnard, Kobus},
  booktitle={Proceedings of the IEEE/CVF Winter Conference on Applications of Computer Vision},
  pages={950--959},
  year={2021}
}

@inproceedings{yu2020searching,
  title={Searching central difference convolutional networks for face anti-spoofing},
  author={Yu, Zitong and Zhao, Chenxu and Wang, Zezheng and Qin, Yunxiao and Su, Zhuo and Li, Xiaobai and Zhou, Feng and Zhao, Guoying},
  booktitle={Proceedings of the IEEE/CVF Conference on Computer Vision and Pattern Recognition},
  pages={5295--5305},
  year={2020}
}

@article{wu2023repisd,
  title={Rep{ISD-N}et: Learning efficient infrared small-target detection network via structural re-parameterization},
  author={Wu, Shuanglin and Xiao, Chao and Wang, Longguang and Wang, Yingqian and Yang, Jungang and An, Wei},
  journal={IEEE Transactions on Geoscience and Remote Sensing},
  volume={61},
  pages={1--12},
  year={2023},
  publisher={IEEE}
}

@article{li2026dynamic,
  title={Dynamic high-frequency convolution for infrared small target detection},
  author={Li, Ruojing and Xiao, Chao and Yin, Qian and An, Wei and Chen, Nuo and Ying, Xinyi and Li, Miao and Wang, Yingqian},
  journal={IEEE Transactions on Circuits and Systems for Video Technology},
  year={2026},
  volume={36},
  number={6},
  pages={7676--7680},
  publisher={IEEE}
}

@article{liu2024infraredb,
  title={Infrared small target detection via joint low rankness and local smoothness prior},
  author={Liu, Pei and Peng, Jiangjun and Wang, Hailin and Hong, Danfeng and Cao, Xiangyong},
  journal={IEEE Transactions on Geoscience and Remote Sensing},
  volume={62},
  pages={1--15},
  year={2024},
  publisher={IEEE}
}

@article{liu2025star,
  title={{STAR-Net}: An interpretable model-aided network for remote sensing image denoising},
  author={Liu, Jingjing and Jin, Jiashun and Xiu, Xianchao and Zhang, Jianhua and Liu, Wanquan},
  journal={Pattern Recognition},
  pages={112496},
  year={2026},
  publisher={Elsevier}
}

@article{zhou2022infrared,
  title={Infrared small target detection via $\ell_0$ sparse gradient regularized tensor spectral support low-rank decomposition},
  author={Zhou, Fei and Fu, Maixia and Duan, Yule and Dai, Yimian and Wu, Yiquan},
  journal={IEEE Transactions on Aerospace and Electronic Systems},
  volume={59},
  number={3},
  pages={2105--2122},
  year={2022},
  publisher={IEEE}
}

@article{yin2026dnn,
  title={{DNN}-aided low-rank and sparse decomposition model for infrared small target detection},
  author={Yin, Jia-Jie and Li, Heng-Chao and Zheng, Yu-Bang and Geng, Xiong-Fei and Pan, Jie},
  journal={Pattern Recognition},
  pages={113070},
  year={2026},
  publisher={Elsevier}
}

@article{hao2026deep,
  title={Deep Unfolding Residual Decomposition for Infrared Small Target Detection},
  author={Hao, Fan and Wei, Feng and Zhou, Feng and Wang, Zhipeng and Yao, Shichao and Yang, Xueyan and Ma, Zongfang},
  journal={IEEE Transactions on Geoscience and Remote Sensing},
  year={2026},
  volume={64},
  number={},
  pages={5006013},
  publisher={IEEE}
}

@article{bai2010analysis,
  title={Analysis of new top-hat transformation and the application for infrared dim small target detection},
  author={Bai, Xiangzhi and Zhou, Fugen},
  journal={Pattern Recognition},
  volume={43},
  number={6},
  pages={2145--2156},
  year={2010},
  publisher={Elsevier}
}

\end{document}